\documentclass[journal]{IEEEtran}

\usepackage{cite}

\usepackage{tabularx}

\usepackage[CJKbookmarks=true]{hyperref}
\newcommand{\reftab}[1]{Table \ref{#1}}

\usepackage{color}

\ifCLASSINFOpdf
\usepackage[pdftex]{graphicx}
\newcommand{\reffig}[1]{Fig. \ref{#1}}
\graphicspath{{image}}
\DeclareGraphicsExtensions{.pdf,.jpeg,.png}
\else

\fi

\usepackage{amsmath}
\usepackage{mathtools}
\usepackage{footmisc}   

\usepackage{amsthm}
\theoremstyle{plain}

\usepackage{amssymb}

\usepackage{multirow}
\usepackage{tabulary}
\usepackage[utf8]{inputenc}

\usepackage{subfigure}

\usepackage{lineno}

\begin{document}


\bstctlcite{IEEEexample:BSTcontrol}

\title{Fast and Accurate Road Crack Detection Based on Adaptive Cost-Sensitive Loss Function}

\author{Kai~Li,~Bo~Wang,~Yingjie~Tian,~and~Zhiquan~Qi

\thanks{This work has been partially supported by grants from: National Natural Science Foundation of China (Nos. 12071458, 71731009, and 61702099) and the Fundamental Research Funds for the Central Universities in UIBE (No.CXTD10-05).}

\thanks{Kai Li is with the School of Mathematics Sciences, University of Chinese Academy of Sciences, Beijing 100049, China. (e-mail: likai14@mails.ucas.ac.cn)}
\thanks{Yingjie Tian and Zhiquan Qi are with the Research Center on Fictitious Economy and Data Science, Chinese Academy of Sciences, Beijing 100190, China; with the Key Laboratory of Big Data Mining and Knowledge Management, Chinese Academy of Sciences, Beijing 100190, China.}

\thanks{Bo Wang is with University of International Business and Economics, Beijing 100029, China. (e-mail: wangbo@uibe.edu.cn)}
\thanks{The first two authors contributed equally to this work.}
\thanks{Corresponding Authors: Yingjie~Tian (e-mail: tyj@ucas.ac.cn), Zhiquan Qi (e-mail: qizhiquan@foxmail.com).}}

\maketitle

\begin{abstract}
Numerous detection problems in computer vision, including road crack detection, suffer from exceedingly foreground-background imbalance. Fortunately, modification of loss function appears to solve this puzzle once and for all. In this paper, we propose a pixel-based adaptive weighted cross-entropy loss in conjunction with Jaccard distance to facilitate high-quality pixel-level road crack detection. Our work profoundly demonstrates the influence of loss functions on detection outcomes, and sheds light on the sophisticated consecutive improvements in the realm of crack detection. Specifically, to verify the effectiveness of the proposed loss, we conduct extensive experiments on four public databases, i.e., CrackForest, AigleRN, Crack360, and BJN260. Compared with the vanilla weighted cross-entropy, the proposed loss significantly speeds up the training process while retaining the performance.
\end{abstract}

\begin{IEEEkeywords}
Crack detection, Jaccard distance, U-Net, weighted cross-entropy (WCE).
\end{IEEEkeywords}

\IEEEpeerreviewmaketitle

\section{Introduction}
\IEEEPARstart{R}{oad} crack detection is critical to pavement quality maintenance and distress prediction for the sake of transportation safety \cite{oliveira2009automatic}.
However, traditional manual road crack detection approaches are extremely time-consuming and labor-intensive with subjective mis-detections \cite{cheng1999novel,cheng2001novel,subirats2006automation,nguyen2009automatic}. Consequently, when automatic detection makes its appearance, it has been appealing to researchers to develop rapid and dependable crack analysis in intelligent transportation systems \cite{oliveira2013automatic,shi2016automatic}.

In terms of crack detection, arduous efforts have been made to yield automatic detection through {deep networks \cite{adu2011multiresolution, fan2018automatic, huang2018deep, jenkins2018deep, cheng2018pixel, ji2018automated, konig2019convolutional, bang2019encoder, 2019Real, 2018NB, Li2020Automatic, fang2019distribution, song2020automated}.} Nonetheless, most of them only focus on the tricks in network design, which is rather subtle and trivial at times. The achievement mainly benefits from \emph{end-to-end} design philosophy, which emphasizes {minimal a priori} representational and computational assumptions, and seeks to avoid explicit structural dependency and {hand-engineering} involving.

\begin{figure}[htbp]
    \centering
    \includegraphics[width=\linewidth]{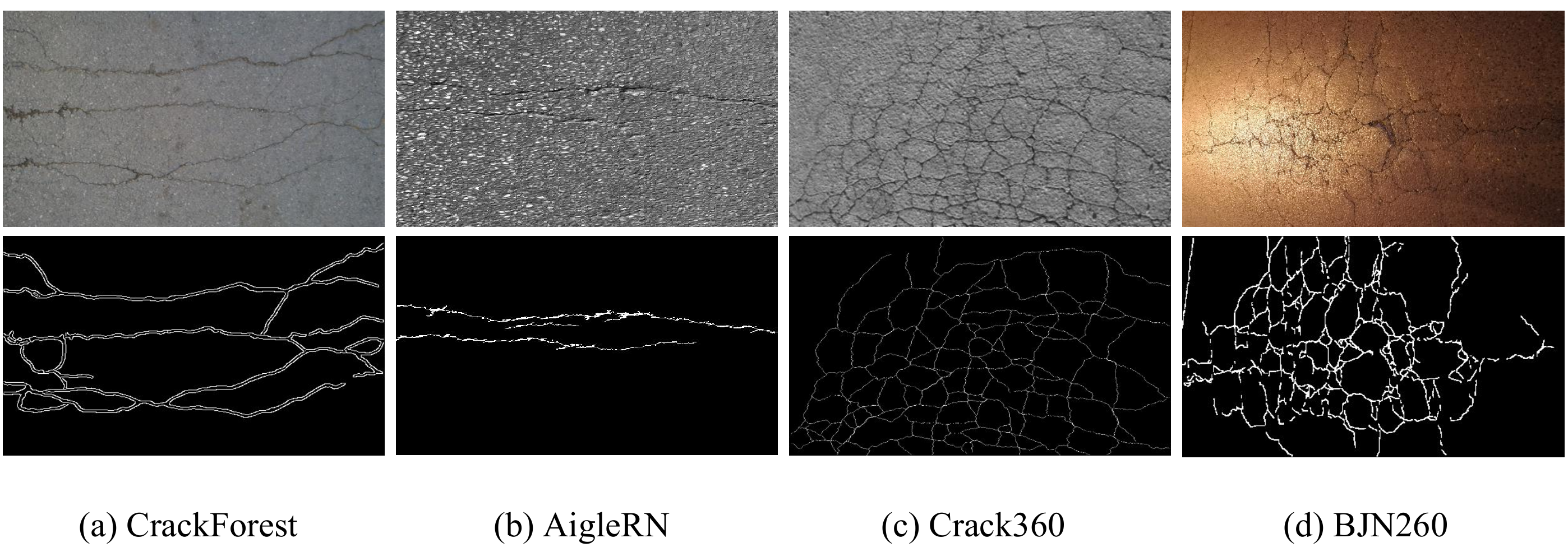}
    \caption{These pictures illustrate the category imbalance problem in crack detection. (a), (b), (c) and (d) show the \emph{densest} crack image (top) and {ground-truth (bottom) from the four training sets, {namely CrackForest \cite{shi2016automatic}, AigleRN \cite{amhaz2015automatic}, Crack360 \cite{zou2018deepcrack}}, {and our BJN260}, respectively.} And the proportion of crack pixels in the single picture is about \emph{3.85\%, 1.13\%, 2.04\%, {and 6.46\%}}, respectively. Besides, the proportion of crack pixels in the four training sets, are approximately about 1.11\%, 0.55\%, 0.53\%, {and 1.56\%}, respectively.}
    \label{fig1}
\end{figure}

Meanwhile, pavement crack detection often encounters the problem with \emph{extreme category imbalance}, that is, crack pixels are far fewer than the non-crack ones (see \reffig{fig1}). Unfortunately, in the field of crack detection, this knowledge in structural priority is rarely touchable by previous methods based on machine learning, thus hardly boosts the ultimate performance.
{Accordingly, we argue that it is the imbalance of different classes in pixel-level that creates the \emph{main} obstacle of fast and accurate road crack detection.}

Typically, there are at least three regimes to resolve category imbalance.
Firstly, {a naive way is to apply ensemble principles\cite{galar2012a, feng2018class, fernandes2019ensemble, zefrehi2020imbalance}. The final decision classifier is usually obtained by several weak classifiers through scoring strategies. Obviously, this approach usually involves a large number of calculation.}
Secondly, another common method is sampling technique \cite{japkowicz2002class,estabrooks2004multiple,liu2019model, yan2019oversampling, tsai2019under}.
However, sampling methods often distort the underlying data distribution. Besides, how to choose the proper sampling rate also becomes an unavoidable issue.

{Then, {the third approach is cost-sensitive learning\cite{sun2007cost-sensitive, khan2018cost-sensitive, Yuan2020SteeringLoss}.} Generally, this method needs to introduce an imbalanced cost matrix of misclassifications. In addition, cost-sensitive learning is an equivalent way to sampling technique if we regard the sampling method as a pipeline that explicitly conveys cost-sensitivity to the appearances of examples. Nevertheless, this system usually has to confront a tough challenge, that is, how to define a suitable cost for the misclassifications.

Based on these considerations, we apply novel adaptive cost-sensitive loss functions to confront the dilemma of imbalance for road crack detection, incorporating a deep end-to-end network in this paper.
{Specifically, our loss functions are motivated by the weighted cross-entropy (WCE) in \cite{xie2015holistically} and \cite{fang2019distribution}. {To balance the loss between major and minor categories {automatically}, they both utilize a ratio as the weight for the minor class. In detail, the ratio refers to the proportion of the number of pixels of ground-truth major class and minor class in a batch of training data. Based on sample statistics to achieve adaptive weight, we focus on studying how to learn the appropriate weight term for the under-represented category, in order to detect road cracks \emph{fast} and \emph{accurately.}}
Note that `fast' in this paper refers to accelerating training a model instead of improving the training or testing speed of the model. In addition, by virtue of the significant capability of Jaccard distance in image segmentation, we also adopt it into our proposed loss functions for crack detection.

In summary, our contributions lie in four-fold.
\begin{itemize}
  \item We propose three novel adaptive WCE losses for dealing with crack detection.
  \item Regarding the penalty $q$ for the minor category in WCE, we find that: a) reducing it properly can speed up the training process greatly and b) it is better not to exceed the specified upper bound, namely, $10$.
  \item {Besides, we collect a pavement crack database in Beijing's night scenes, BJN260. It is shared to the community to facilitate crack detection research. To our knowledge, it is the first road crack dataset for night scenes.}
  \item Finally, compared with the vanilla weighted cross-entropy \cite{xie2015holistically}, our methods could markedly shorten the training time while retaining test accuracy.
%
\end{itemize}

The remainder of this paper is organized as follows. Section \ref{part2} reviews the related work. Section \ref{part3} explores our novel adaptive losses for crack detection. Section \ref{part4} demonstrates the effectiveness of our systems by experiments. Finally, Section \ref{part5} takes further discussions by combining our loss with other state-of-the-art models.

\section{Related Work}\label{part2}
In this section, we begin with crack detection methods based on shallow models and then introduce some approaches based on deep learning.
Finally, we discuss cost-sensitive learning for the imbalanced data.

\subsection{Shallow Models for Crack Detection}
The traditional {shallow models for} crack detection can be categorized into several communities.

Firstly, wavelet transform based on multi-scale analysis can be applied to cracks and non-cracks separation \cite{subirats2006automation}. It takes advantage of continuous two-dimensional wavelet transform to build an elaborate coefficient graph, but fails to handle cracks with low continuity and high curvature, due to the anisotropic characteristic of wavelet.

Secondly, the contrast ratio is used to detect the visually conspicuous salient regions from the surroundings \cite{achanta2008salient}. However, it hardly yields results with satisfying completeness and continuity \cite{arbelaez2011contour}.

Thirdly, the texture-analysis method pays attention to the texture feature of the pavement distress with local information on every pixel \cite{hu2010novel}. This method is performed in pixel-to-pixel fashion and vastly loses neighbor information, resulting in neglecting the cracks with intensity in-homogeneity.

Besides, minimal path selection seems to be promising in detecting contour-like image structures, as well as the width of the crack \cite{kass1988snakes, kaul2012detecting, amhaz2014new, Amhaz2016Automatic}. Nevertheless, when it comes to real-life applications, this method turns out to be infeasible because of the intensive computation cost.

Last but not least, in the realm of traditional machine learning, the primary principle spotlights how to separate crack pixels from the background. On this occasion, previous researches, such as \cite{oliveira2013automatic, shi2016automatic, strisciuglio2017detection} and \cite{strisciuglio2019robust}, mostly focus on shallow models with handcrafted feature extraction or feature engineering as well, which may vary among different situations, resulting in reduced overall performance.

\subsection{Deep Convolutional Neural Networks for Crack Detection}
Compared with the above traditional methods, there has been a recent wave of development \cite{adu2011multiresolution, zhang2016road, huang2018deep, cha2017deep, fan2018automatic, ji2018automated, jenkins2018deep, cheng2018pixel, konig2019convolutional, bang2019encoder, 2019Real, song2020automated, 2018NB, Li2020Automatic, wu2019face, he2019asymptotic, Yanan2020automatical} using deep convolutional neural networks {(DCNNs)} that emphasizes the importance of  \emph{automatic} feature learning.

According to the modeling mechanism, we divide the pixel-level crack detection methods based on deep learning into two groups as follow.

a) \emph{Methods Based on Classification Networks:} For example, Zhang et al.\cite{zhang2016road} cropped training images into small patches, and then utilize DCNNs to automatically learn features and detect road cracks in pixel-level.
Fan et al.\cite{fan2018automatic} considered crack detection as a multi-label classification problem through extracting small pieces from crack images as inputs to generate extensive training data.


In general, the appropriate image patches play an essential role in these systems. Specifically, one has to address the following issues: how to generate patches, how to choose the size of them, and how many patches to generate.

b) \emph{Approaches Based on Fully Convolutional Networks:}
These methods usually utilize skip layers to fuse low-level and high-level semantic information.
{For example, Yang et al.\cite{yang2018automatic} applied fully convolutional networks \cite{long2015fully} to fulfill automatic crack detection.
Li et al.\cite{Li2020Automatic} applied U-Net \cite{ronneberger2015u} and a CNN with alternately updated clique \cite{yang2018convolutional} to fulfil automatic crack detection. Fang et al.\cite{fang2019distribution} utilized U-Net with a weighted cross-entropy loss for road crack detection.
Based on Deeplab v3+ \cite{2018Encoder}, Song et al. \cite{song2020automated} proposed a novel network, CrackSeg, to capture rich multi-scale features. Zou et al.\cite{yang2018automatic} utilized hierarchical convolutional features to achieve accurate crack detection by combining SegNet \cite{badrinarayanan2017segnet} and a holistically-nested edge detection algorithm \cite{xie2015holistically}. Besides, the proposed edge detection method by Liu et al. \cite{liu2017richer} is often used as a solution of crack detection.}

Compared with the first methods, {the second ones directly implement the crack detection task in an image-to-image manner and requires no data preprocessing before mathematical modeling}. For the crack detection task, in this paper we focus on the adaptive loss functions dealing with category imbalance instead of designing a novel architecture. {In other words, our work is orthogonal and complementary with those researches of designing network architectures.}

\subsection{Cost-sensitive Learning for Imbalanced Data}
Cost-sensitive learning is a longstanding topic and has received much attention \cite{maloof2003learning,mccarthy2005does,liu2006influence}. As a common approach to address with imbalanced data, non-uniform cost offers an alternative strategy to the conventional sampling methods. In detail, besides introducing the cost matrix (putting higher cost for misclassification on the minor class), the threshold of the Bayesian classifier could be altered by base rate modification on a particular class \cite{elkan2001foundations}.

Hence, cost-sensitive learning {is helpful for learning from imbalanced} data. In other words, we can modify the cost metric to achieve the same effect as that of the sampling methods. An obvious merit of the cost-sensitive approach is that it can eliminate the downside occurred by sampling mentioned above.

Accordingly, a straightforward tool is an adaptive loss function that can offset the overwhelming dominant class for the minority. In terms of loss function revision, the methods based on weighted cross-entropy (WCE) are often appealing to previous researches, especially in an encoder-decoder model for detection. For example, Xie et al.\cite{xie2015holistically} utilized the ratio of ground-truth edge and non-edge in a batch of training data to reweight non-edge and edge samples in WCE loss, respectively. {Besides, Fang et al.\cite{fang2019distribution} took advantage of similar adaptive WCE to deal with category imbalance.} In this paper, we mainly consider how to reweight the under-represented category and then propose three novel WCE loss functions.

\section{Methodology}\label{part3}
In this section, we begin with the network for crack detection and then introduce some fundamental work in loss functions. Next we discuss the effective way of cost-sensitive loss design for imbalanced pixel-level classification and then propose our loss functions for crack detection.

\subsection{Network Architecture for Crack Detection}
Inspired by \cite{cheng2018pixel} and \cite{ji2018automated}, we directly employ U-Net \cite{ronneberger2015u} as our network with minor revisions\footnote{The main difference is that we choose \emph{same padding} in our network instead of valid padding in the original U-Net, to keep the size of feature maps identity. Besides, instead of up-convolution in U-Net, we choose \emph{deconvolution} for the up-sampling stages.}. The architecture is illustrated in \reffig{our U-Net architecture}.

\begin{figure}[ht]
  \centering
  \includegraphics[width=8.8cm]{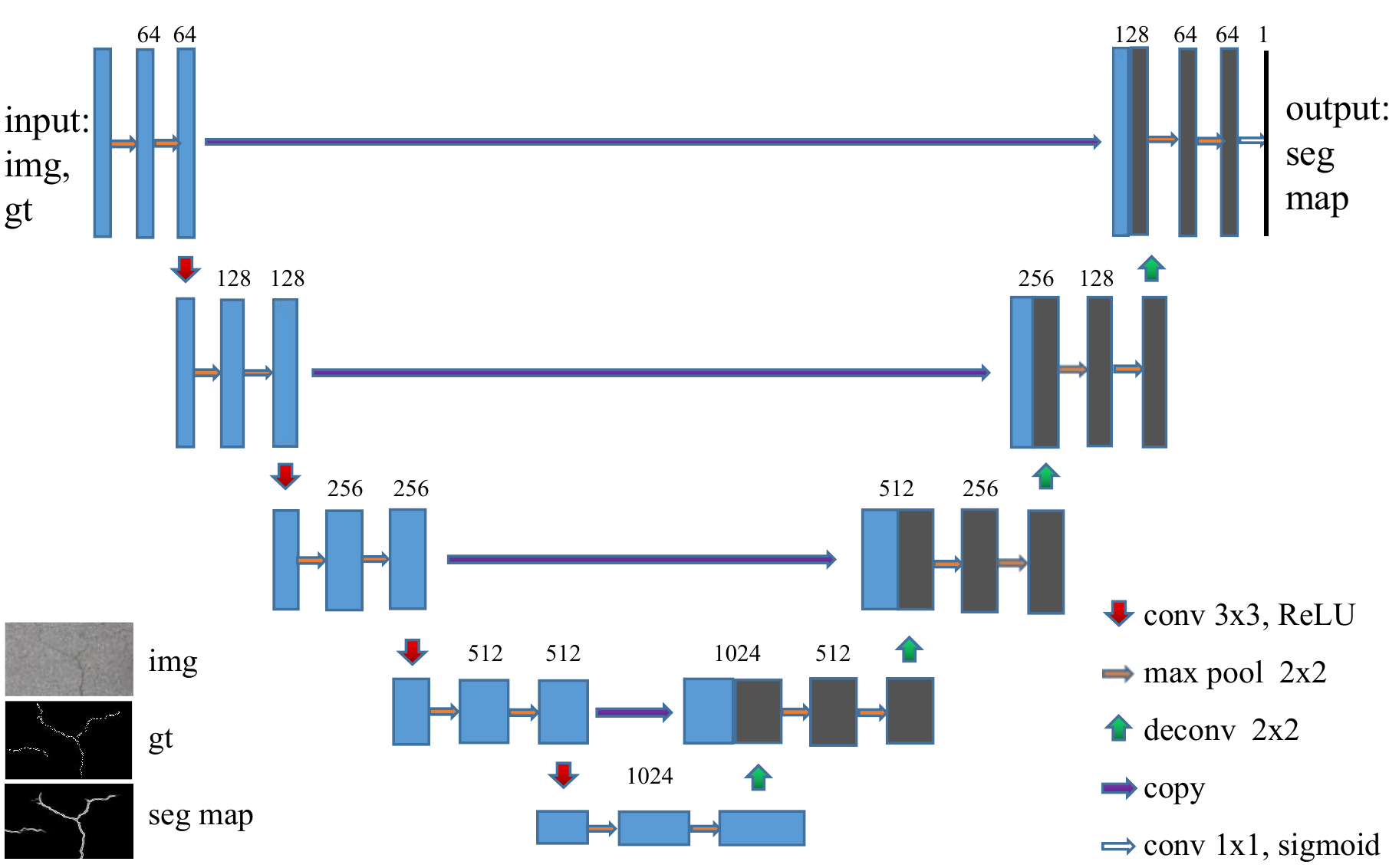}\\
  \caption{The overall architecture of U-Net. Note that `img', `gt', and `seg map' refer to `image', `ground-truth', and `segmentation map', respectively. The digit above every rectangle indicates the channel number of the corresponding feature map. }
 \label{our U-Net architecture}
\end{figure}

\subsection{Sensitivity of Weights in Crack Detection}\label{Sensitivity of Weights in Crack Detection}

\begin{figure*}[htbp]
    \centering
    \includegraphics[height=6cm]{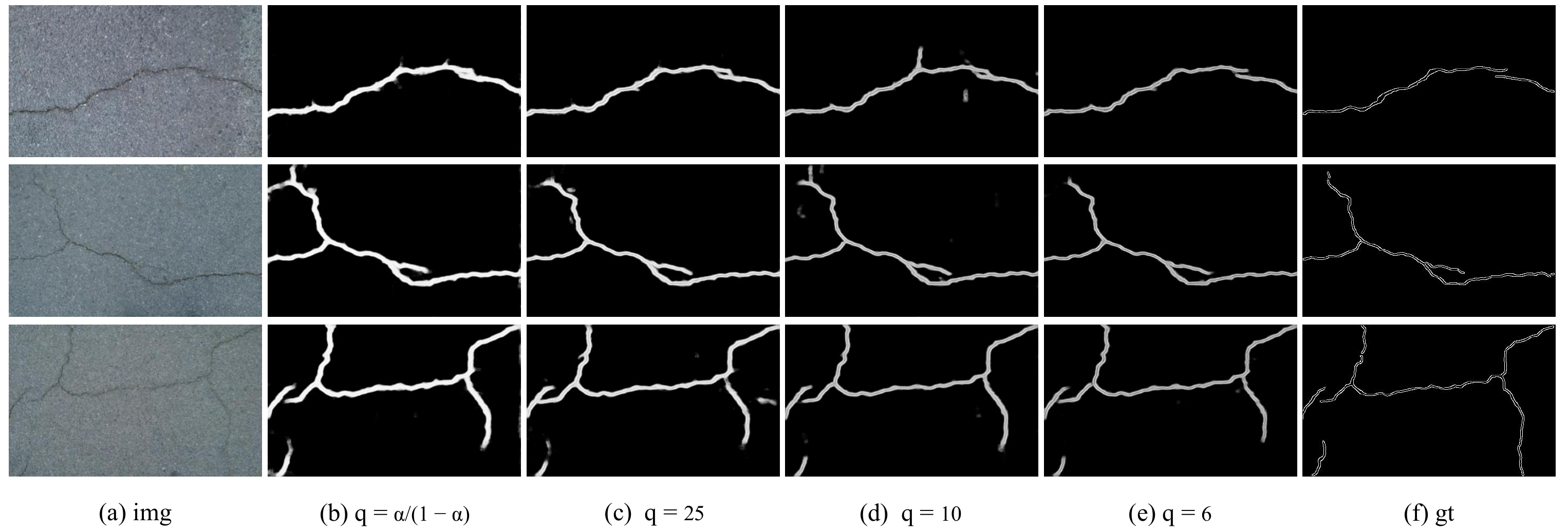}
    \caption{The sensitive analysis of crack detection results concerning different penalty coefficients. (a) contains the raw images from CrackForest; (b$\sim$e) are the results obtained by training U-Net under different penalties in \eqref{eq1}; (f) is the corresponding {ground truth}. Note that $\alpha$ is the proportion of non-crack pixels in a batch of training data. Here, $25$ is about the minimum integer ratio of non-crack v.s. crack pixels in a single ground-truth image. In other words, $\alpha /(1-\alpha)\ge 25$.}
    \label{fig3}
\end{figure*}

The cross-entropy loss is often used in many deep learning models for classification. However, it fails to attain an acceptable recall rate with the imbalanced data.
Hence, many efforts have been made in handling this imbalanced situation. For instance, for the image edge detection task, to deal with the imbalance between the edge and non-edge pixels, Xie et al. propose the following WCE \cite{xie2015holistically}:
\begin{equation}\label{eq}
    \small
    L(\mathbf{y}, \mathbf{p})=-\sum_{j}\big[\alpha \, y_j \, log \, p_j + (1-\alpha)\,(1-y_j) \, log(1-p_j)\big],
\end{equation}
where $\alpha$ refers to the ratio of {ground-truth} non-edge pixels in a batch of training data. Besides, $y_j$ and $p_j$ mean the real label and the posterior possibility of the $j$-th pixel. This method thereby becomes an important example of {how to achieve a relatively high recall} rate while keeping a reasonable prediction precision.

Obviously, \eqref{eq} is equivalent to the following formula:
\begin{equation}\label{eq1}
    \small
    L(\mathbf{y}, \mathbf{p})=-\sum_{j}\big[q(\alpha) \, y_j \, log \, p_j + (1-y_j) \, log(1-p_j)\big],
\end{equation}
where
\begin{equation}\label{q_xie}
    \small
    q(\alpha) \!= \alpha/(1-\alpha).
\end{equation}
{Furthermore, for a supervised classification problem, \eqref{eq1} utilizes the statistical information of samples to achieve adaptive weight for the minor category in the loss function.}

Nevertheless, using $q(\alpha)$ in \eqref{q_xie} as the overall penalty for the minor class is sometimes \emph{ineffective} in specific applications, such as crack detection.
As shown in \reffig{fig3}, the results manifest that the relative frequency is not a suitable choice for extremely imbalanced data. In particular, in the realm of crack detection, it misclassifies the dominant class into the minor category, i.e., outputting a \emph{{thicker}} detection structure than the ground-truth. Besides, through \reffig{fig3}, we explicitly illustrate the effects of other penalties. One may find easily that the crack detection results lose fidelity when the penalty $q(\alpha)$ for the minor class is increasing.

\begin{figure*}[htbp]
    \centering
    \subfigure[Batch size {=} 1]{
    \includegraphics[width=4.2cm]{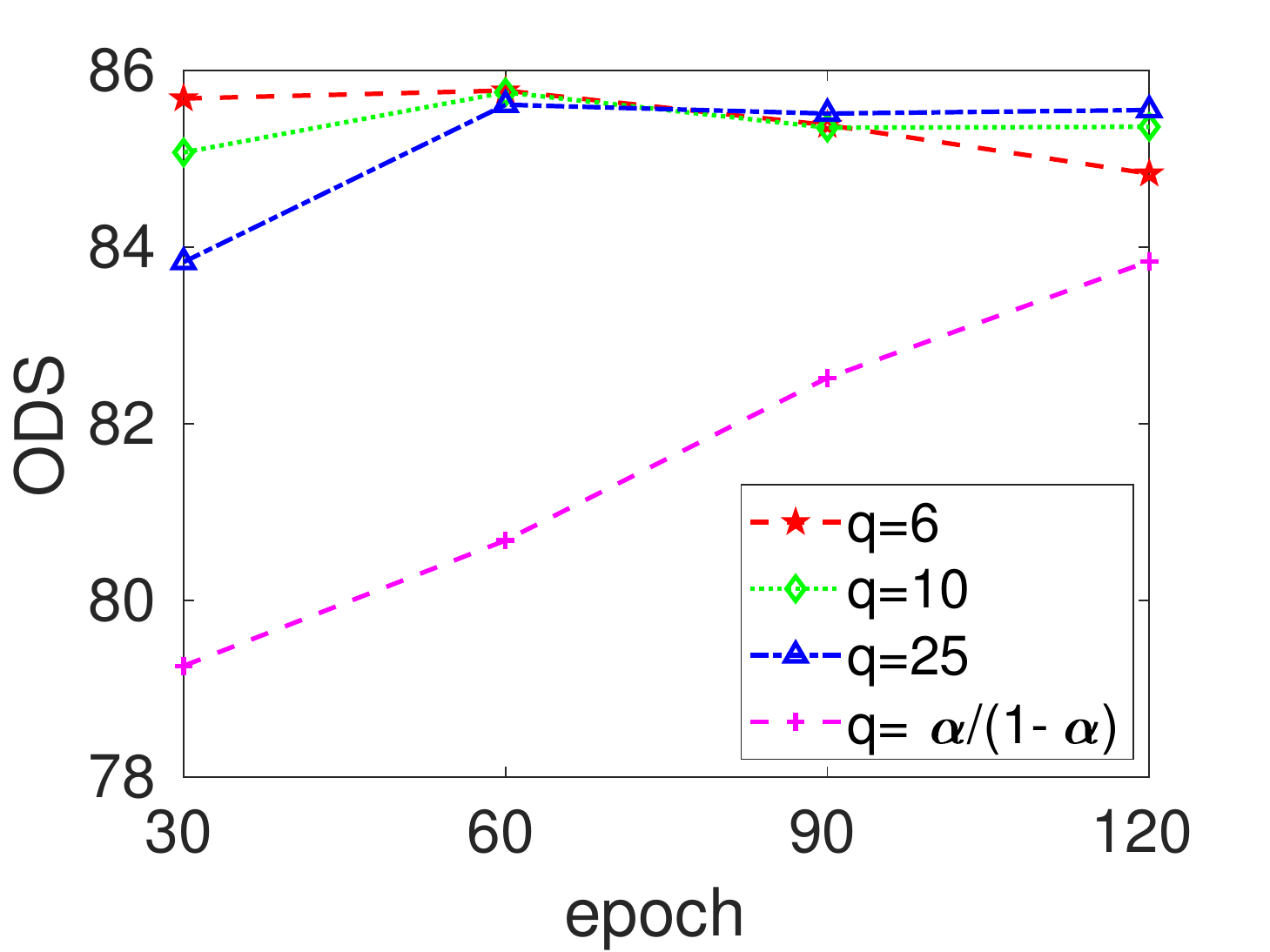}
    }
    \subfigure[Batch size {=} 2]{
    \includegraphics[width=4.2cm]{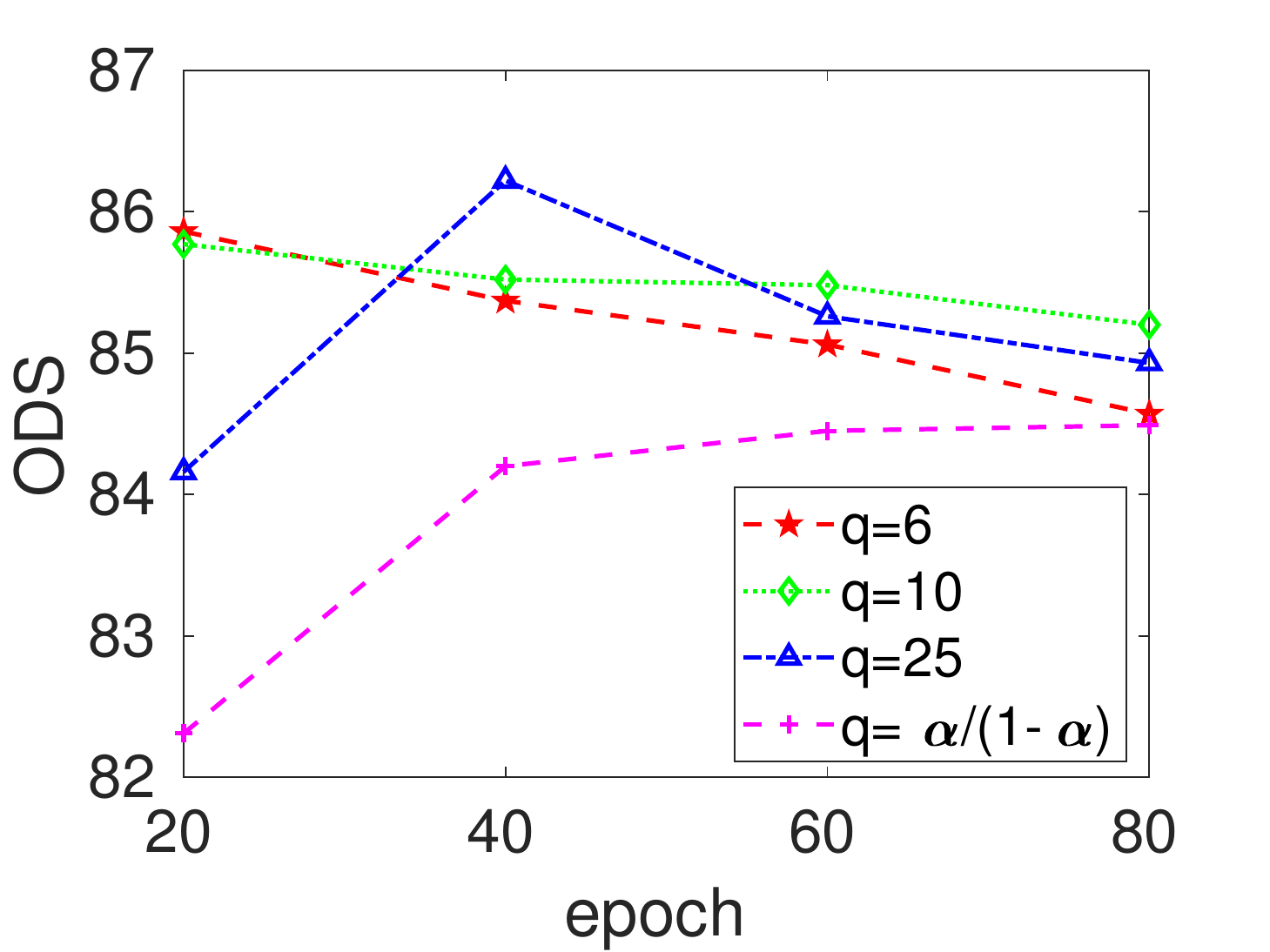}
    }
    \subfigure[Batch size {=} 4]{
    \includegraphics[width=4.2cm]{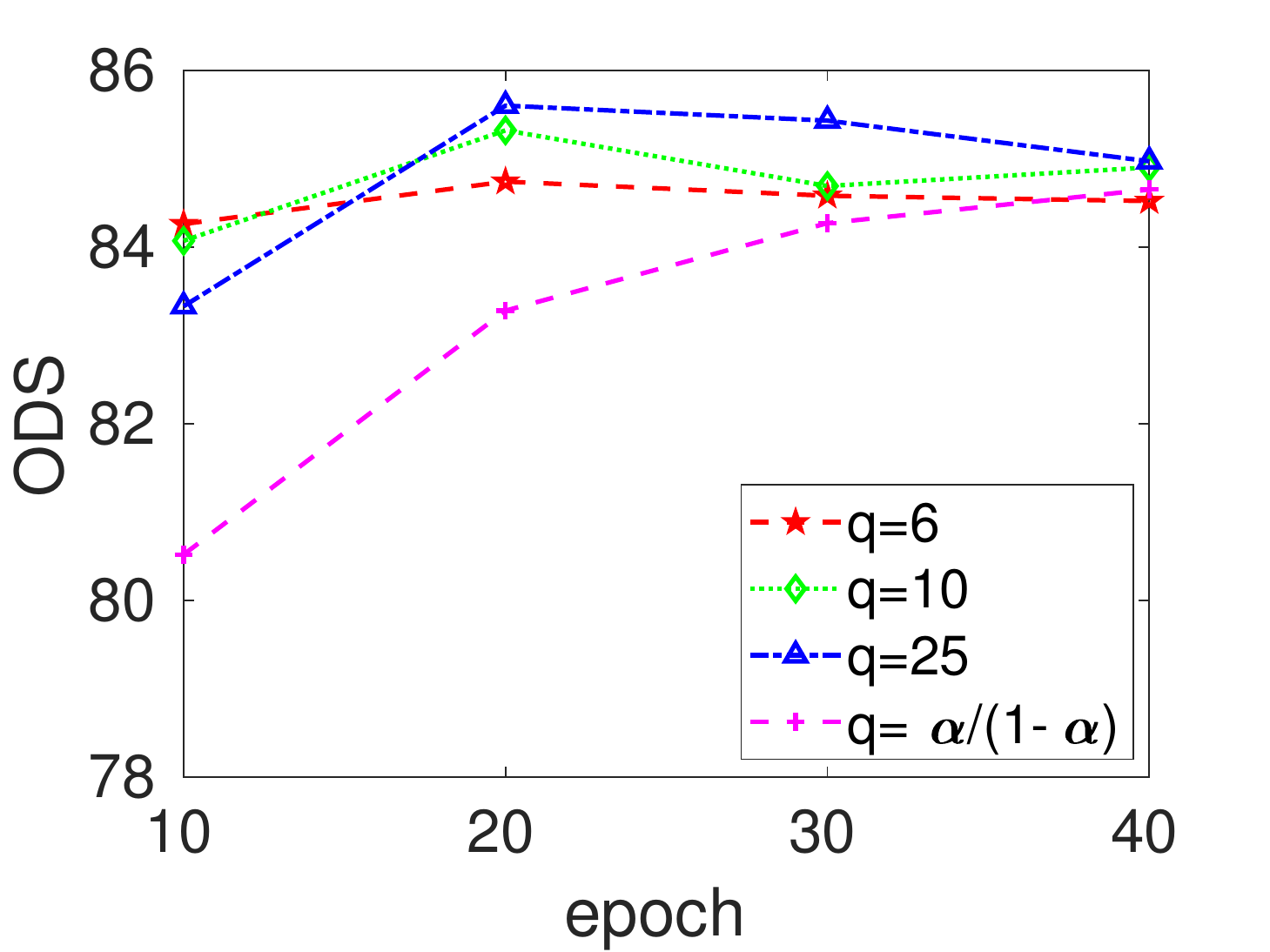}
    }
    \subfigure[Batch size {=} 8]{
    \includegraphics[width=4.2cm]{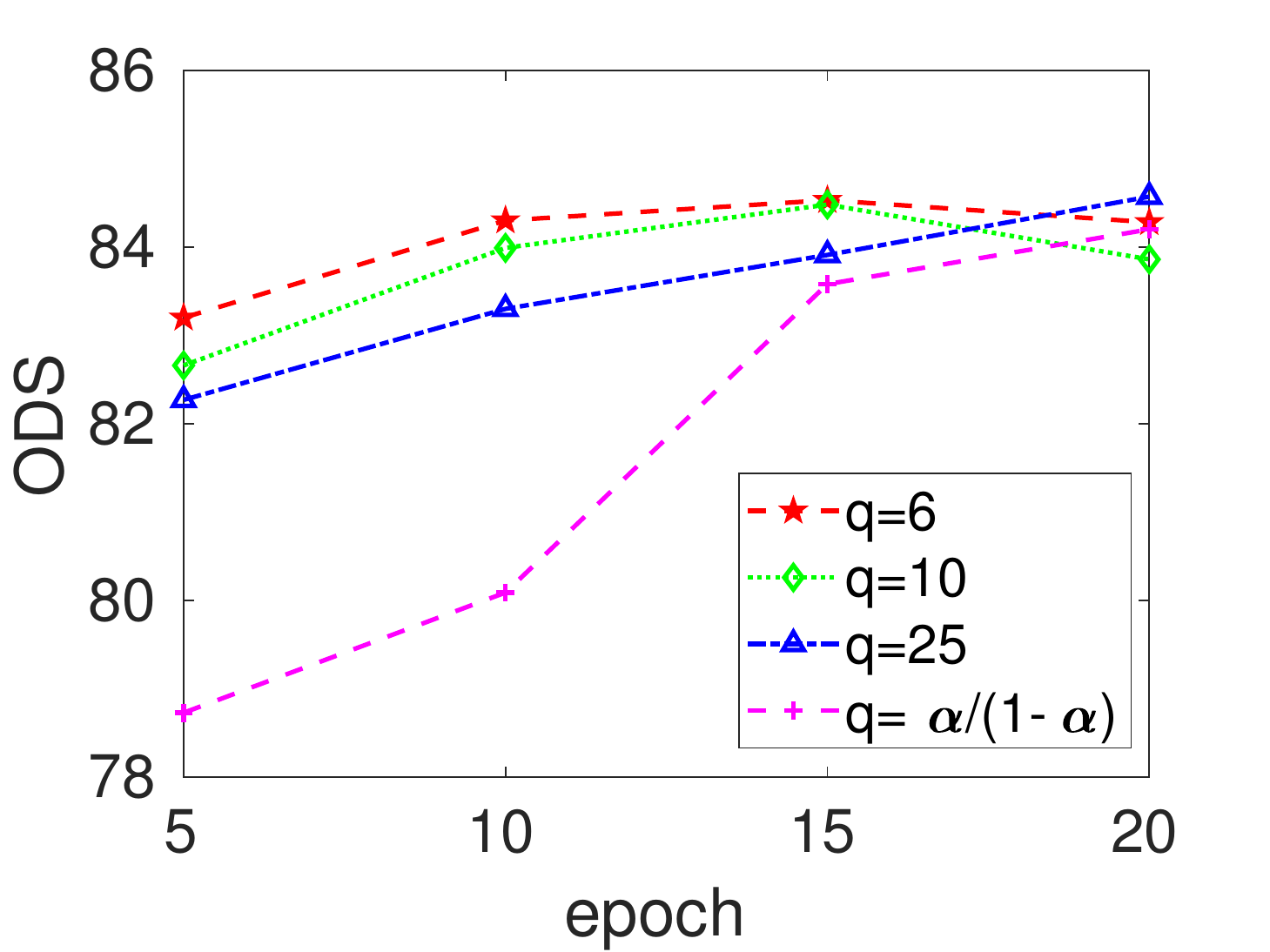}
    }
    \caption{Performance on CrackForest under different penalty coefficients. The horizontal coordinate `epoch' indicates the training epoch of U-Net; the vertical coordinate `ODS' indicates F{$_1$} score based on the optimal dataset scale.}
    \label{Performance under different penalty hyperparameters}
\end{figure*}

On the other hand, different penalties result in different training time, although the training time of single epoch is invariant for the same network.  Through \reffig{Performance under different penalty hyperparameters}, we illustrate some results in terms of ODS (which is the F{$_1$} score based on optimal dataset scale) under different penalties with different batch sizes. One may find that reducing the penalty for the crack category \emph{properly} could speed up the training process while obtaining a good performance. {Thus, based on sample statistical information to achieve adaptive weight, this paper aims to research how to learn the appropriate penalty term for the minor category.}
}

\subsection{The Proposed Weighted Cross-entropy Families}
As discussed in \ref{Sensitivity of Weights in Crack Detection}, the key of fast and accurate crack detection is {properly} decreasing $q(\alpha)$ in \eqref{eq1}. To this end, we can leverage several function families to guide this adjustment.

\subsubsection{\textbf{Power function type}}

Inspired by focal loss \cite{lin2018focal} resolving category imbalance, we propose the power weighted method:
 \begin{equation}\label{Power function type}
   q(\alpha)=\beta * \big(\frac{\alpha}{1-\alpha}\big)^\gamma, 0<\beta,\gamma \leq 1.
 \end{equation}
Here, ${\beta}$ is a hyperparameter used to fine-tune. A simple strategy for it is to utilize quartiles or octaves.
Note that there are two differences\footnote{They mean that: (a) Priori information is utilized instead of the posterior used in object detection. In terms of supervised classification, one could know beforehand whether each pixel in images is a positive sample or not.
(b) In addition, to accelerate training the model, the penalty \emph{q} needs to be reduced rather than increased for the minor class, as shown in {\reffig{fig3}} and \reffig{Performance under different penalty hyperparameters}.} between the proposed power function and that of focal loss \cite{lin2018focal}.

\subsubsection{\textbf{Logarithmic function type}}
Besides the square root operation, another common way to reduce a numerical value is the logarithmic operation. Thus, we propose the following weighted method:
\begin{equation}\label{Logarithmic function type}
    q(\alpha)=\beta * ln\big(\frac{\alpha}{1-\alpha}\big), 0<\beta \leq 1.
\end{equation}
Besides, one may need Laplacian smoothing for the two methods above when the denominator approaches zero.

\subsubsection{\textbf{Exponential function type}}
Besides the quotient between sample proportion in a batch of training data, one could utilize their difference to deal with the category imbalance. To widen the gap between sample ratios and make the penalty $q(\alpha)>1$, we introduce the exponential weighted method:
\begin{equation}\label{Exponential function}
    q(\alpha)=\beta * a^{\gamma(2\alpha-1)}, 0<\beta \leq 1, a>1, 0 \leq \gamma \leq 1.
    \end{equation}
Considering the training time and the evaluation metric via \reffig{Performance under different penalty hyperparameters}, we use $a=10$ and $\gamma=1$ in practice:
\begin{equation}\label{Exponential function type}
    q(\alpha)=\beta * 10^{2\alpha-1}, 0<\beta \leq 1.
\end{equation}
Obviously, $ q(\alpha) \leq 10 $ in \eqref{Exponential function type}, because $0<\beta \leq 1, 0<\alpha \leq 1$.

\subsection{{Holistic Loss Function for} Crack Detection}
Besides weighted cross-entropy, we also utilize Jaccard-index \cite{jaccard1912distribution} (denoted as ${d}_J$) to measure the differences between images. By combining them, we formulate the holistic loss:
\begin{equation}
    Holistic(\mathbb{Y}, \mathbb{\tilde{Y}})= a * L(\mathbb{Y}, \mathbb{\tilde{Y}})+ b * {d}(\mathbb{Y}, \mathbb{\tilde{Y}}).
    \label{the lotal loss}
\end{equation}
Here, ${\mathbb{Y}}$ and $\mathbb{\tilde{Y}}$ refer to the ground-truth image set and the prediction image set in a batch of training data, respectively. Besides, $a$ and $b$ are the hyper-parameters for the trade-offs,
\begin{align}
    {d}(\mathbb{Y}, \mathbb{\tilde{Y}})  &= 1 - {d}_J(\mathbb{Y}, \mathbb{\tilde{Y}}),\\
    {d}_J(\mathbb{Y}, \mathbb{\tilde{Y}}) &= \frac{\sum_{j=1}^{N} {y}_j \cdot {p}_j + \lambda}  {\sum_{j=1}^{N} {y}_j + \sum_{j=1}^{N} {p}_j \, -  \sum_{j=1}^{N}{{y}_j \cdot {p}_j} +\lambda},
    \label{eq9}
\end{align}
where $\lambda$ is an additional Laplace smoothing item.

\section{Experiments}\label{part4}
All experiments in this paper are conducted using a GTX 1080Ti with CPU E5-2680v2.
We implement our network using the publicly available Keras 2.1.0 based on TensorFlow \cite{abadi2016tensorflow}. Specifically, we evaluate the proposed {approaches} on four public datasets: CrackForest \cite{shi2016automatic}, AigleRN \cite{2011Automatic}, Crack360{\cite{zou2018deepcrack}}, and our BJN260.

In this paper, we employ two baseline models, i.e., 1) random structured forest (RSF)\cite{shi2016automatic} and 2) WCE \cite{xie2015holistically} combined with the revised U-Net. Unless otherwise specified, during training U-Net, we utilize the following data augmentation and hyper-parameter configurations. Besides, some metrics are introduced for evaluation.

{\textbf{Data augmentation}}:
 We apply similar data augmentation {to the implementation{\footnote{https://github.com/zhixuhao/unet}, which refers to the rotation, shifting, zoom, flipping, shear, and padding with neighboring pixels.

{\textbf{Hyper-parameter setting}}:
During training, we utilize full images as the input with batch size 2. The channels in the down-sampling stages are $[64,128,256,512,1024]$, respectively. The Adam optimizer \cite{kingma2015adam} is adopted to train U-Net (300 steps every epoch), where the initial learning rate is empirically set as $3\times10^{-4}$. For the holistic loss in \eqref{the lotal loss}, we first consider the effect of different WCE losses by setting $a=1$ and $b=0$, and then fine-tune ${a}$ by fixing {$b=1$}.

{\textbf{Evaluation}}:
{Three metrics are employed, including precision (\textbf P), recall (\textbf R), and F${_1}$ score (\textbf{F${_1}$}), defined as
\begin{equation}\label{precision}
  \textbf{P} = \frac{TP}{TP + FP}, \,\,\,
  \textbf{R} = \frac{TP}{TP + FN}, \,\,\,
  \textbf{F${_1}$} = \frac{2  \textbf{P}  \textbf{R}}{\textbf{P} + \textbf{R}} .
\end{equation}
where TP, FP, and FN refer to true positive, false positive, and false negative, respectively.}


When evaluating the similarity between the prediction image and its ground-truth, it refers to two different threshold methods: optimal dataset scale (ODS) and optimal image scale (OIS). ODS means that a fixed threshold is set across the whole dataset while OIS describes that the best threshold is chosen for each image. In the experiments, we report the results based on ODS and OIS, respectively.

As shuffling the data during training, we adopt the average results of \emph{five trials} as the final evaluation. All test results are presented in percentage systems.

\begin{figure*}[htbp]
\centering
\subfigure[Power function: $\gamma=1/2$]{
\centering
\includegraphics[width=4.2cm]{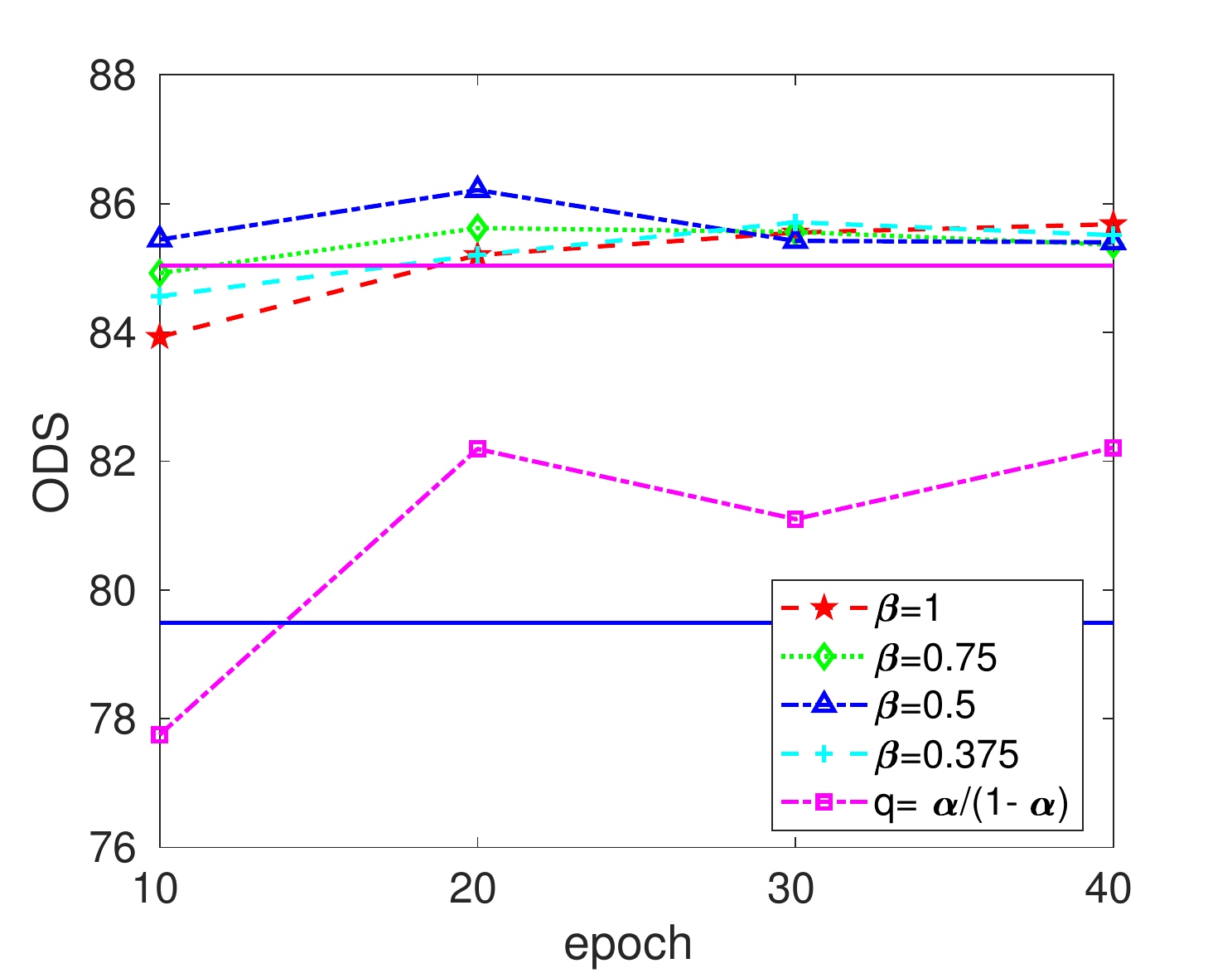}
\label{5a}
}
\subfigure[Power function: $\gamma=1/3$]{
\centering
\includegraphics[width=4.2cm]{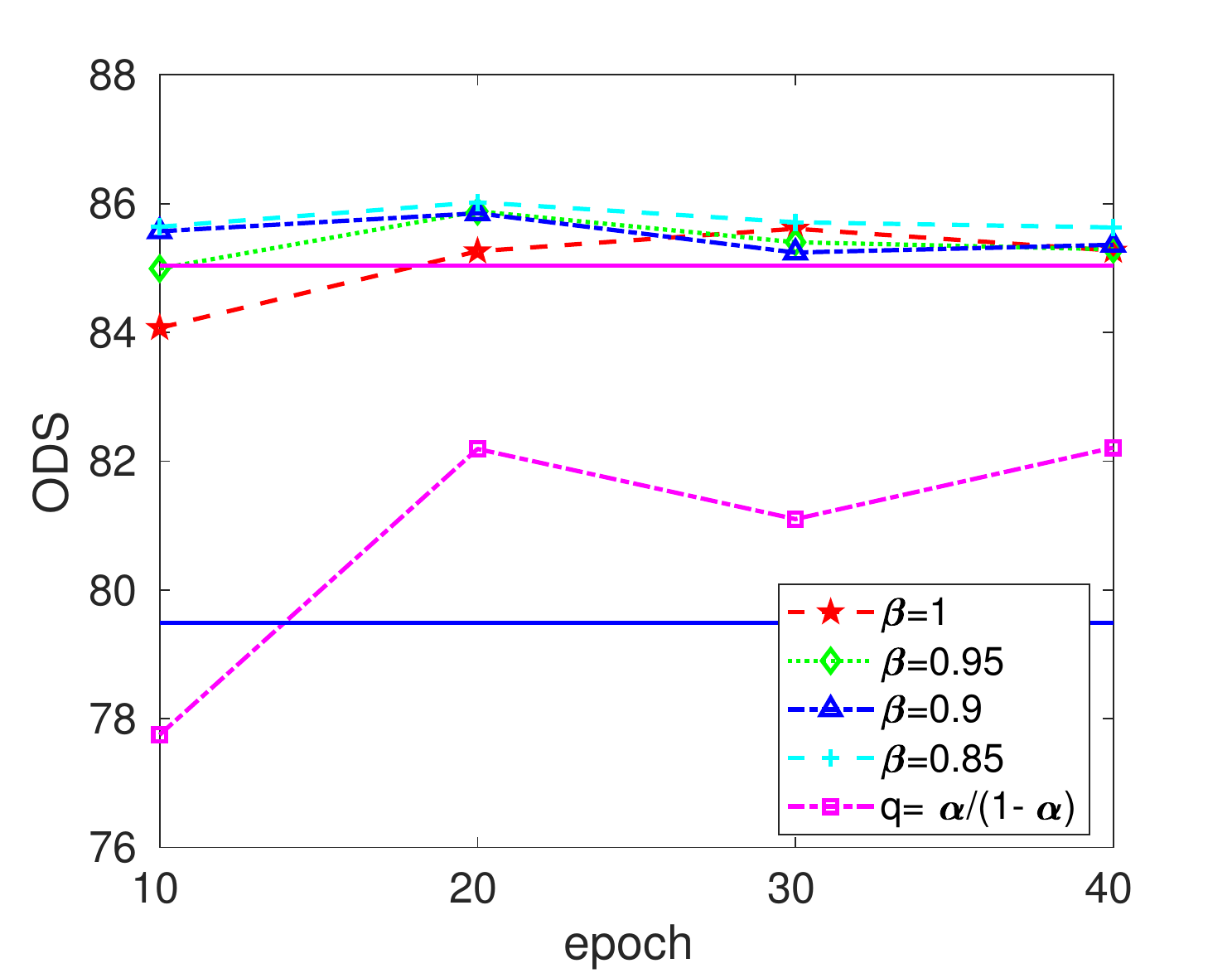}
\label{5b}
}
\subfigure[Logarithmic function]{
\centering
\includegraphics[height=3.3cm, width=4.2cm]{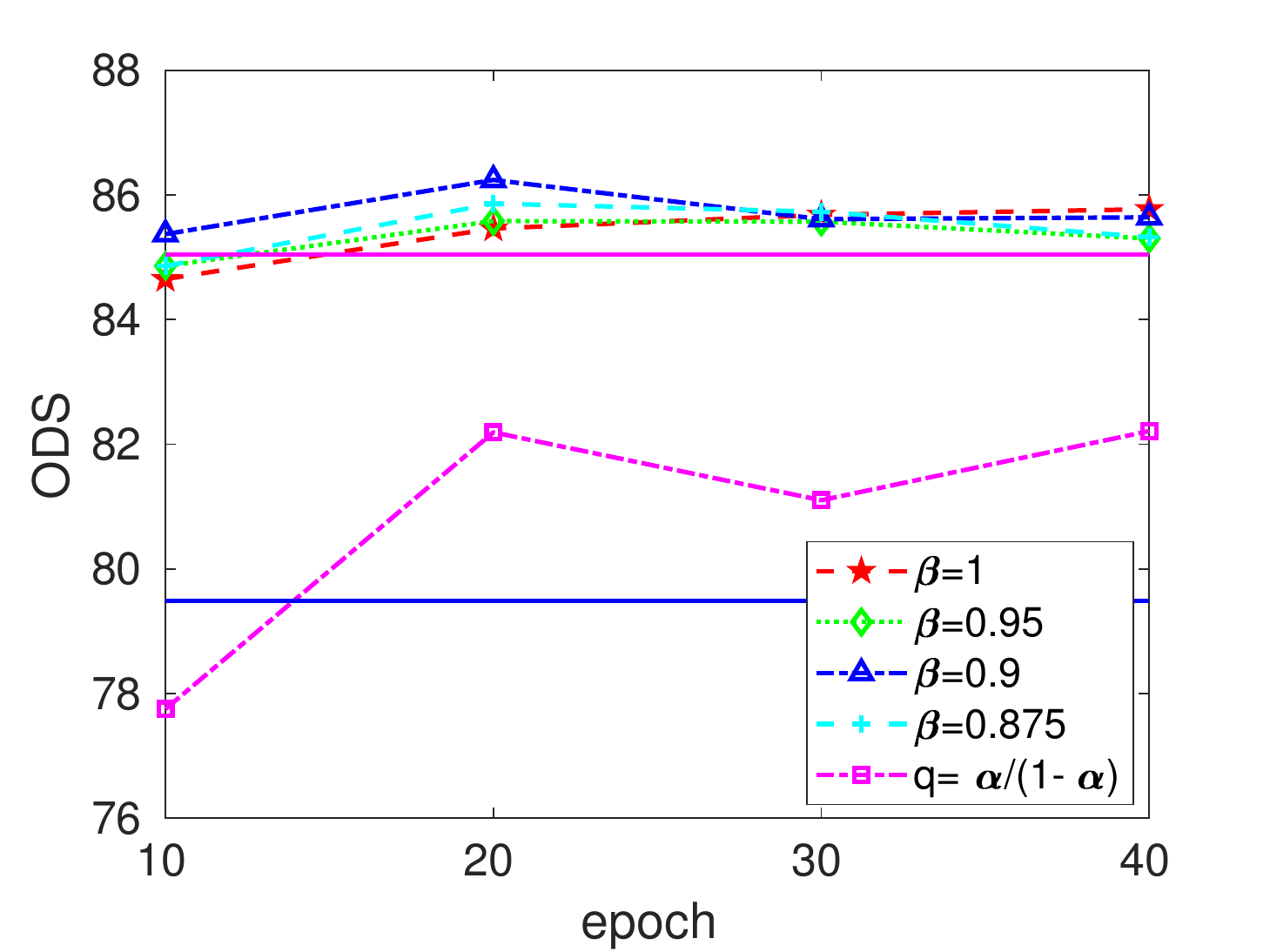}
}
\subfigure[Exponential function]{
\centering
\includegraphics[width=4.2cm]{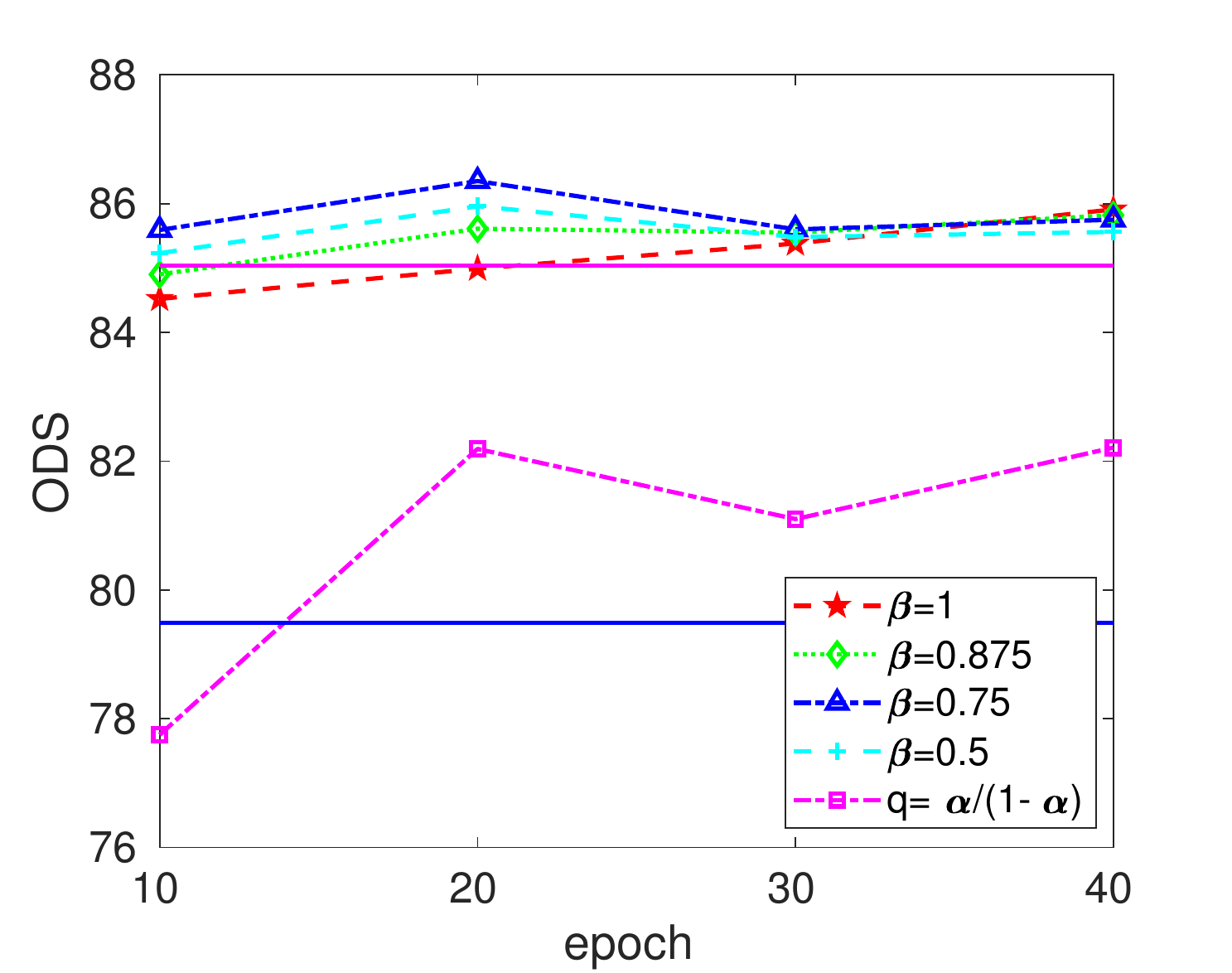}
}
\centering
\caption{The performance under different weighted methods with fine-tuned coefficient $\beta$. The vertical coordinate `ODS' indicates F{$_1$} score based on the optimal dataset scale. The horizontal coordinate `epoch' indicates the training epoch. Besides, the solid blue line shows the first baseline corresponding to the non-deep learning method, random structured forest (RSF)\cite{shi2016automatic}. The solid pink line represents the second baseline training U-Net 70 epochs.}
\label{fig5}
\end{figure*}

\begin{figure*}[htbp]
\centering
\subfigure[Power function : $\gamma=1/2$]{
\centering
\includegraphics[width=4.2cm]{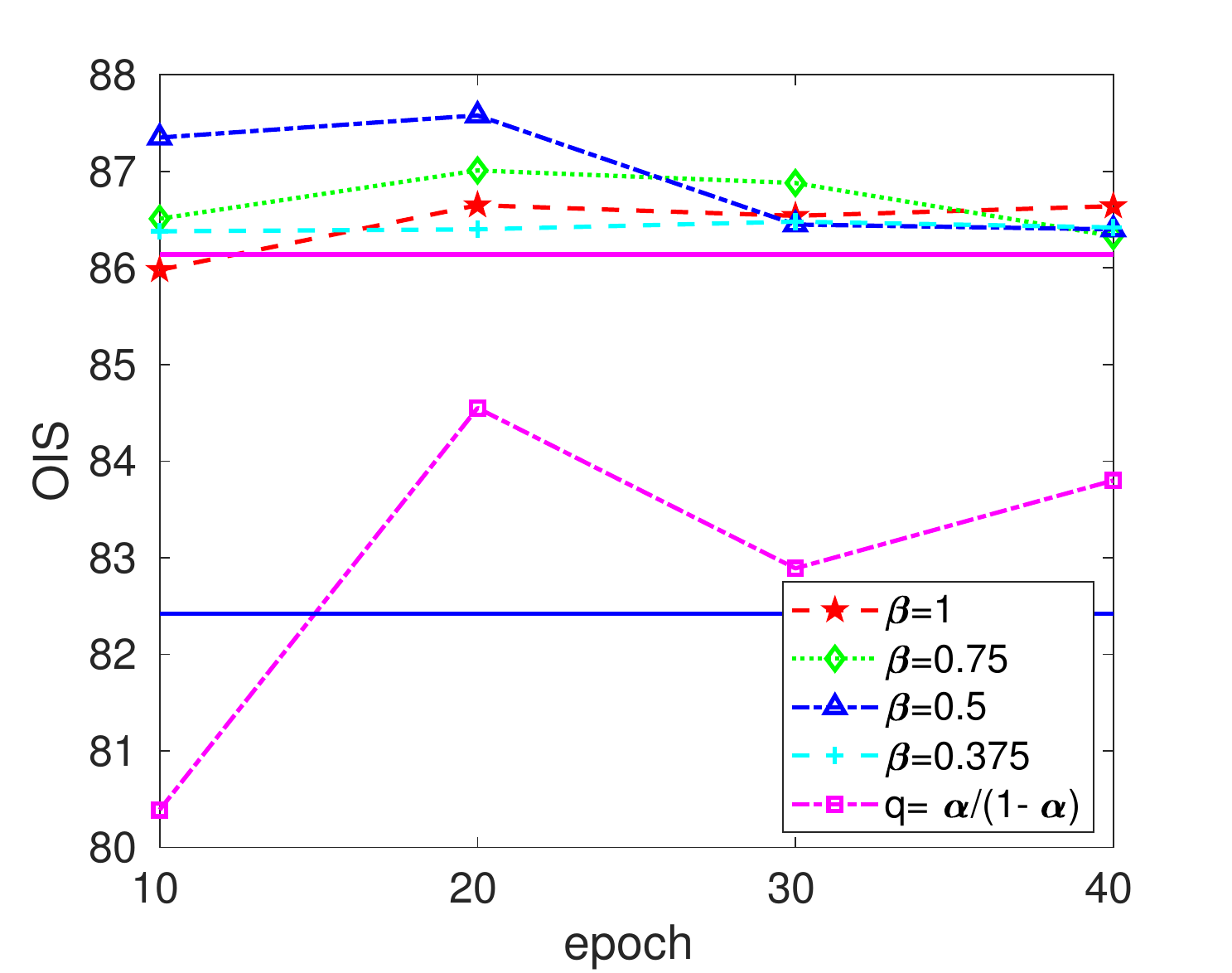}
\label{6a}
}
\subfigure[Power function: $\gamma=1/3$]{
\centering
\includegraphics[width=4.2cm]{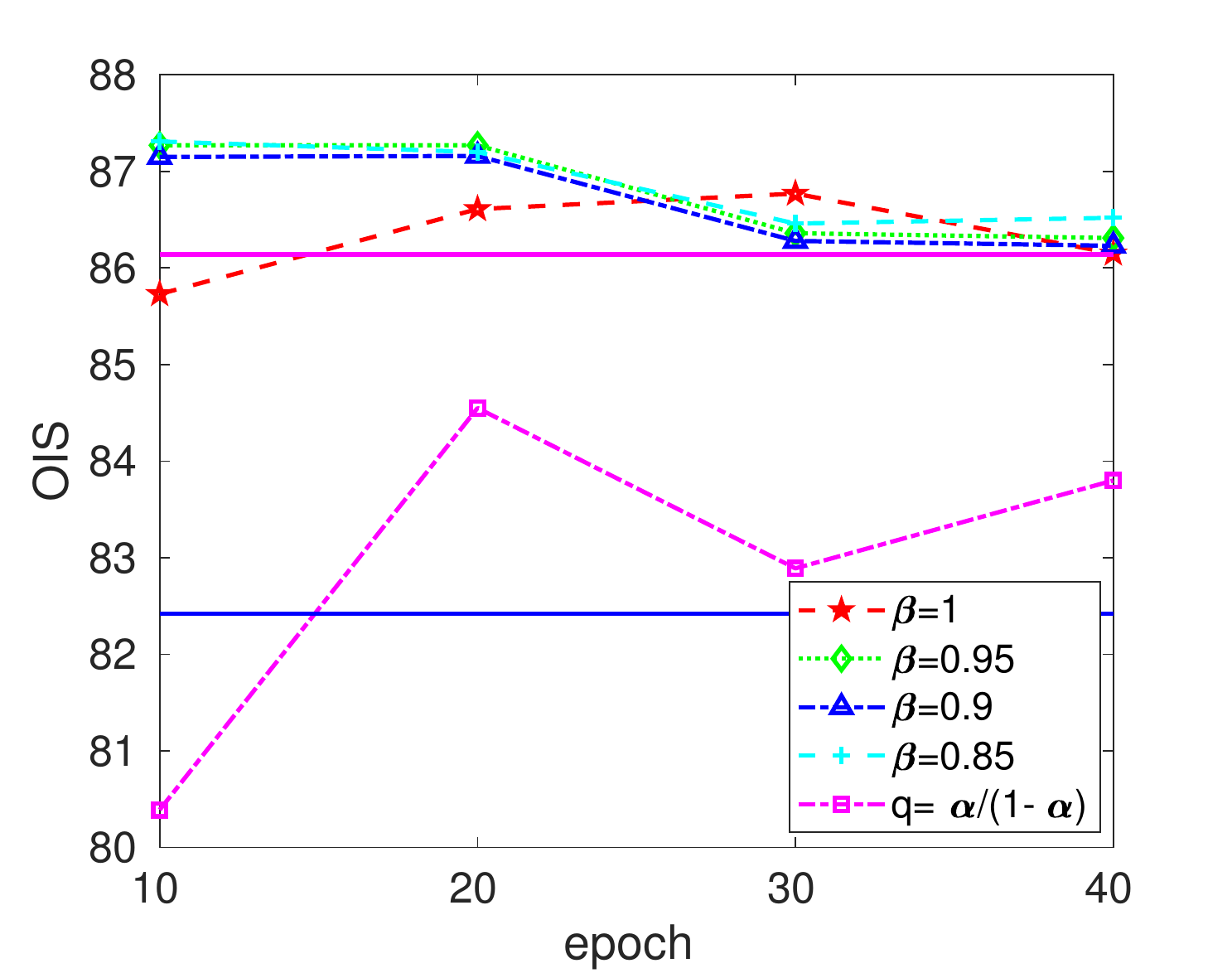}
\label{6b}
}
\subfigure[Logarithmic function]{
\centering
\includegraphics[width=4.2cm]{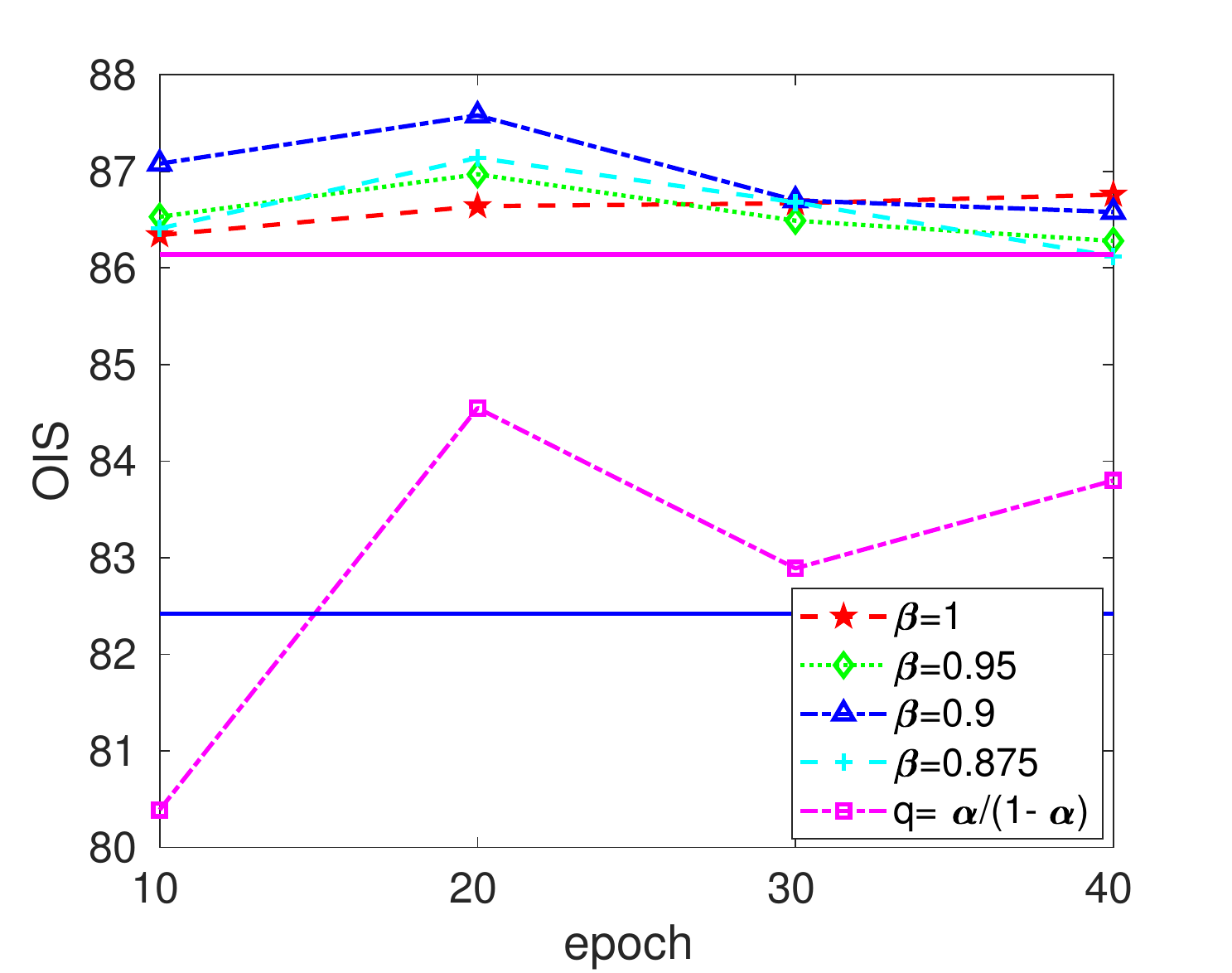}
}
\subfigure[Exponential function]{
\centering
\includegraphics[width=4.2cm]{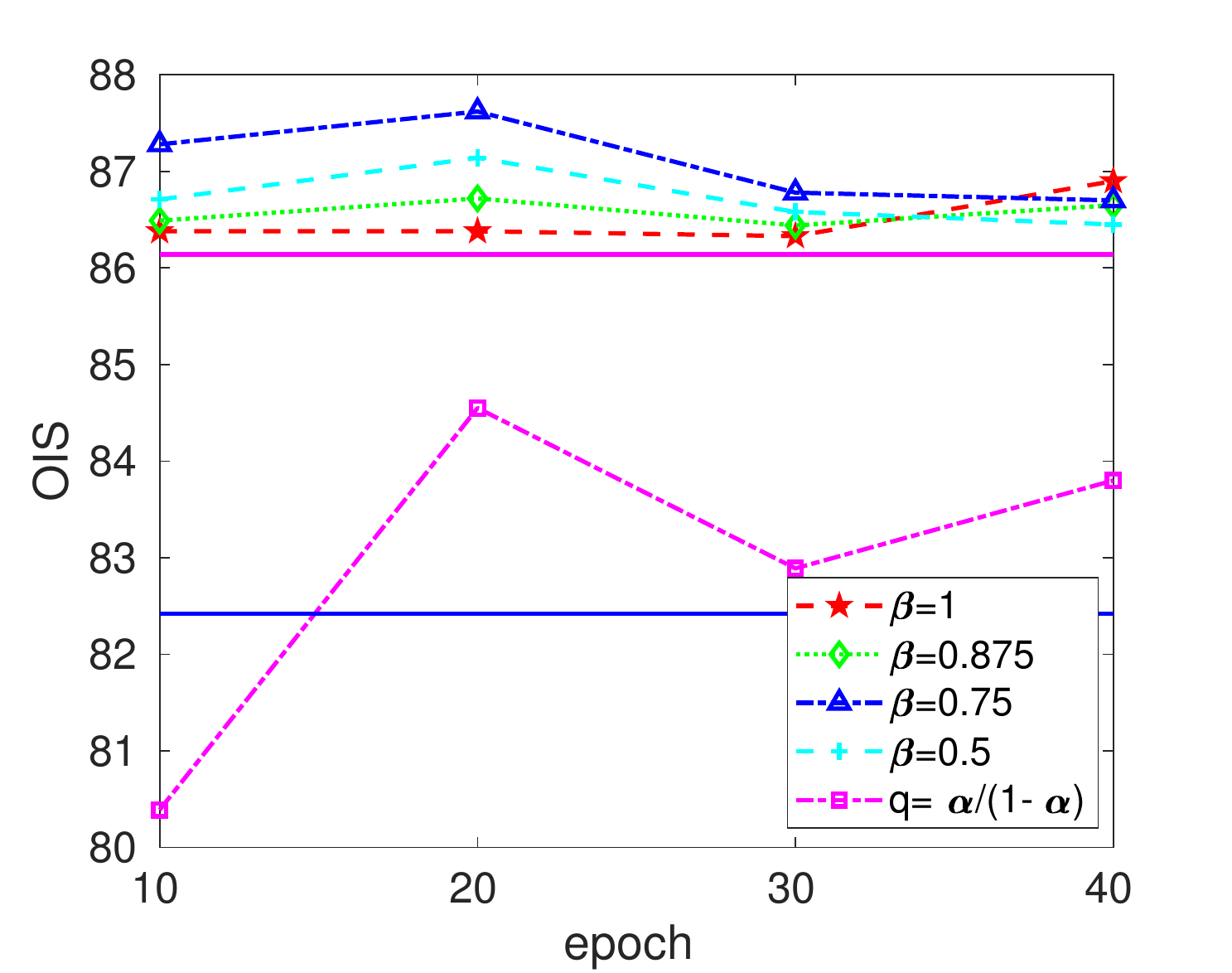}
}
\centering
\caption{The performance under different weighted methods with fine-tuned coefficient $\beta$. The vertical coordinate `OIS' indicates F{$_1$} score based on the optimal image scale. The horizontal coordinate `epoch' indicates the training epoch. Besides, the solid pink line and the solid blue line have the same meaning as those in \reffig{fig5}, respectively.}
\label{fig6}
\end{figure*}

\subsection{Performance {on CrackForest Database}}\label{4A}
There are 118 images in CrackForest \cite{shi2016automatic}, which shares the same size of 480 and 320 pixels in width and height, respectively. Here we divide 60\% and 40\% {of this database into training and testing,  respectively.} For the second baseline, we utilize WCE \eqref{eq1} with \eqref{q_xie} to train U-Net \emph{70} epochs. Except for the weight $q(\alpha)$ in \eqref{eq1} and the training epoch, we utilize the same hyper-parameter setting with the second baseline.

\subsubsection{\textbf{Results of different WCE losses}}
Now, we look at the performance of various WCE losses, i.e., power, logarithmic, and exponential weighted types.

According to different weighted methods, we tune $\beta$ and $\gamma$ (if necessary)\footnote{{Here we combine quartiles and some octaves to fine-tune the hyper-parameter ${\beta}$. Let us take the power function type as an example. When applying ${\gamma}=1/2$ to formula \eqref{Power function type}, we first select the quartiles to obtain a rough range of ${\beta}$, such as (0.25,0.5], then utilize the octave in the interval, namely 0.375, to conduct the experiment, as shown in \reffig{5a} and \reffig{6a}. Note that when using a smaller $\beta$, such as 0.25, the training loss appears non phenomena, so we do not display the corresponding experimental results. A similar phenomenon appears for the power weighted method when ${\gamma}=1/3$ with $\beta=$ 0.75 and 0.8. So we make a further attempt at each interval of 0.05, i.e., utilize $\beta$=0.85, 0.9, 0.95, and 1 to conduct our experiments respectively, as shown in \reffig{5b} and \reffig{6b}.}\label{footnote_4}},
and then obtain the relevant results showed in \reffig{fig5} and \reffig{fig6}. One may find that:


(a) Our weighted methods (fixing $\beta$=$1$) can achieve fast and accurate road crack detection.
For example, the metrics ODS and OIS obtained by 20 training epochs of our methods are better than those by 70 epochs of the second benchmark.

(b) The weight $q(\alpha)$ in \eqref{eq1} for the minor class is not as \emph{small} as possible if fulfilling fast and accurate detection. For example, we choose optimal $\beta$=$0.5$ instead of $\beta$=$0.375$ when using the power function with $\gamma$=$1/2$.

In short, reducing the weight $q(\alpha)$ properly could accelerate training the model while retaining generalization performance. To make a specific comparison, we show some results through \reftab{fast and accurate results on the first database} and then draw the following conclusions:

(a) Apart from \emph{fast} and \emph{accurate}, our proposed {methods} also achieve \emph{stable} crack detection.

In particular, \emph{fast} indicates that our methods utilize 10 training epochs to achieve {at least as good results as} the second baseline using 70 periods. Besides, \emph{accurate} means that our {methods generate at least an 1\%} improvement in F{$_1$} score with 20 training epochs, compared with the second baseline using 70 periods. In addition, \emph {stable} claims that the standard deviations of the five trials using our methods are relatively close to those of the second baseline. Especially, the standard deviation of F{$_1$} score {becomes smaller}.

(b) In terms of F{$_1$} score, the WCE in exponential function has a \emph{slight} advantage over those in power and logarithmic functions under the same training time, such as 20 training epochs.

\begin{table*}[htbp]\footnotesize
    \caption{The average results {with 5 runs} obtained by different WCE losses on CrackForest. Note that `$0.75\_exp\_0.5\_wj$' means applying formula \eqref{the lotal loss} by setting $a=0.5$ and $b=1$ with exponential WCE ($\beta=0.75$).}
    \centering
    \begin{tabular*}{\hsize}{@{}@{\extracolsep{\fill}}c|c|c|c|c|ccc|ccc@{}}
    \hline
    \multirow{2}{*}{Methods}  & \multirow{2}{*}{$\beta$}  & \multirow{2}{*}{$\gamma$}  & \multirow{2}{*}{epoch} & \multirow{2}{*}{time}
    & \multicolumn{3}{c|}{ODS}  & \multicolumn{3}{c}{OIS} \\
    \cline{6-11}
        & & & &             & P            & R            & F{$_1$}                    & P             & R                   & F{$_1$}  \\
    \hline
    RSF \cite{shi2016automatic}
    & - & -  & -  & -       & 87.75        & 72.65        & 79.49         & 89.04   & 76.73          & 82.42  \\
    \hline
    \multirow{4}{*}{$wce\_xie$ \cite{xie2015holistically}}
    & - & -  &70  &54min 54s  & {97.07} (0.22) & 75.66 (1.21) & {85.04} (0.72) & {97.2} (0.16)  & 77.35 (1.14)    & {86.14} (0.66)  \\
    &  &   &20  &15min 44s  & 97.36 (0.44) & 71.12 (1.56) & 82.19 (0.90)  & 97.26 (0.39)  & 74.80 (1.65)    & 84.55 (0.95) \\
    &  &  &10  &7min 54s & 98.12 (0.35) & 64.49 (3.95) & 77.75 (2.82)  & \textbf{97.53} (0.42)  & 68.46 (3.82)    & 80.39 (2.58) \\
    &  &   &5   &3min 39s   & \textbf{98.36} (0.46) & 57.53 (4.69) & 72.49 (3.68)  & {97.37} (0.43)  & 63.17 (4.68)    & 76.53 (3.47)  \\
    \hline
    \multirow{2}{*}{$ wce\_power $}  & \multirow{2}{*}{0.5}  & \multirow{2}{*}{1/2}	
          &20 &15min 44s  &96.59 (0.34)    &77.85 (0.79)	&{86.21} (0.40)           &96.66 (0.34)    &80.07 (0.79)	   &{87.58} (0.36) \\
    & &   &\textbf{10} &\textbf{7min 54s} &96.52 (0.60)    &76.66 (1.47)	&{85.44} (0.68)           &96.73 (0.47)	   &79.64 (1.23)	   &{87.35} (0.58)	\\
    \hline
    \multirow{2}{*}{$ wce\_power $} & \multirow{2}{*}{0.85} & \multirow{2}{*}{1/3}
        &20 &15min 44s &96.15 (1.07)	&77.84 (1.27)	&{86.02} (0.45)           &96.37 (0.88)	   &79.63 (1.13)	   &{87.20} (0.42)	\\
    & & &\textbf{10} &\textbf{7min 54s}  &96.62 (0.23)	&76.62 (0.36)	&{85.47} (0.28)           &96.71 (0.28)	   &79.24 (0.50)	   &{87.11} (0.26)  \\	
    \hline
    \multirow{2}{*}{$ wce\_log $}	& \multirow{2}{*}{0.9} & \multirow{2}{*}{-}
         & 20 &15min 44s    &96.55 (0.22)	&77.93 (1.25)   &{86.24} (0.70)	       &96.66 (0.26)        &80.08 (1.20)	&{87.58} (0.63) \\
    & &  &\textbf{10} &\textbf{7min 54s}   &96.30 (0.86)	&76.72 (2.88)	&{85.37} (1.51)	       &96.52 (0.72)	    &79.36 (2.72)	&{87.08} (1.41) \\
    \hline
    \multirow{3}{*}{$ wce\_exp $}	& \multirow{3}{*}{0.75} & \multirow{3}{*}{1}
         &20 &15min 44s   &96.22 (0.73)	&\textbf{78.34} (1.39)	&{86.35} (0.61)           &96.52 (0.55)	    &\textbf{80.23} (1.15)	&\textbf{87.62} (0.48) \\
    & &  &10 &7min 54s  &96.62 (0.37)	&76.83 (0.83)	&{85.59} (0.37)           &96.84 (0.36)	    &79.44 (0.81)	&{87.28} (0.37) \\
    & &  &5  &3min 39s    &96.46 (0.40)	&76.30 (1.53)	&{85.19} (0.83)	         &96.93 (0.40)	    &78.53 (1.52)	&{86.75} (0.80)	\\
    \hline
    {$ 0.75\_exp\_0.5\_wj$} &0.75	&1 &10	&7min 54s 	&96.62 (0.18)	&78.18 (0.83)	&\textbf{86.42} (0.48)	&96.95 (0.18)   &79.87 (0.84)		&{87.58} (0.46)	\\
    {$ 0.75\_exp\_20\_wj $}	&0.75	&1 &\textbf{5}	&\textbf{3min 39s} 	&96.19 (0.46)	&76.97 (1.04)	&{85.51} (0.56)	&96.48 (0.48)   &79.43 (1.14)		&{87.12} (0.56)	\\
    {$ 0.75\_exp\_50\_wj $}	&0.75	&1 &\textbf{5}	&\textbf{3min 39s} 	&95.74 (1.25)	&77.34 (1.77)	&{85.54} (0.70)	&96.16 (1.12)   &79.66 (1.72)		&{87.12} (0.68)	\\
    \hline
    \end{tabular*}
    \label{fast and accurate results on the first database}
\end{table*}

\begin{figure*}[htbp]
    \centering
    \subfigure{
    \includegraphics[width=4.3cm]{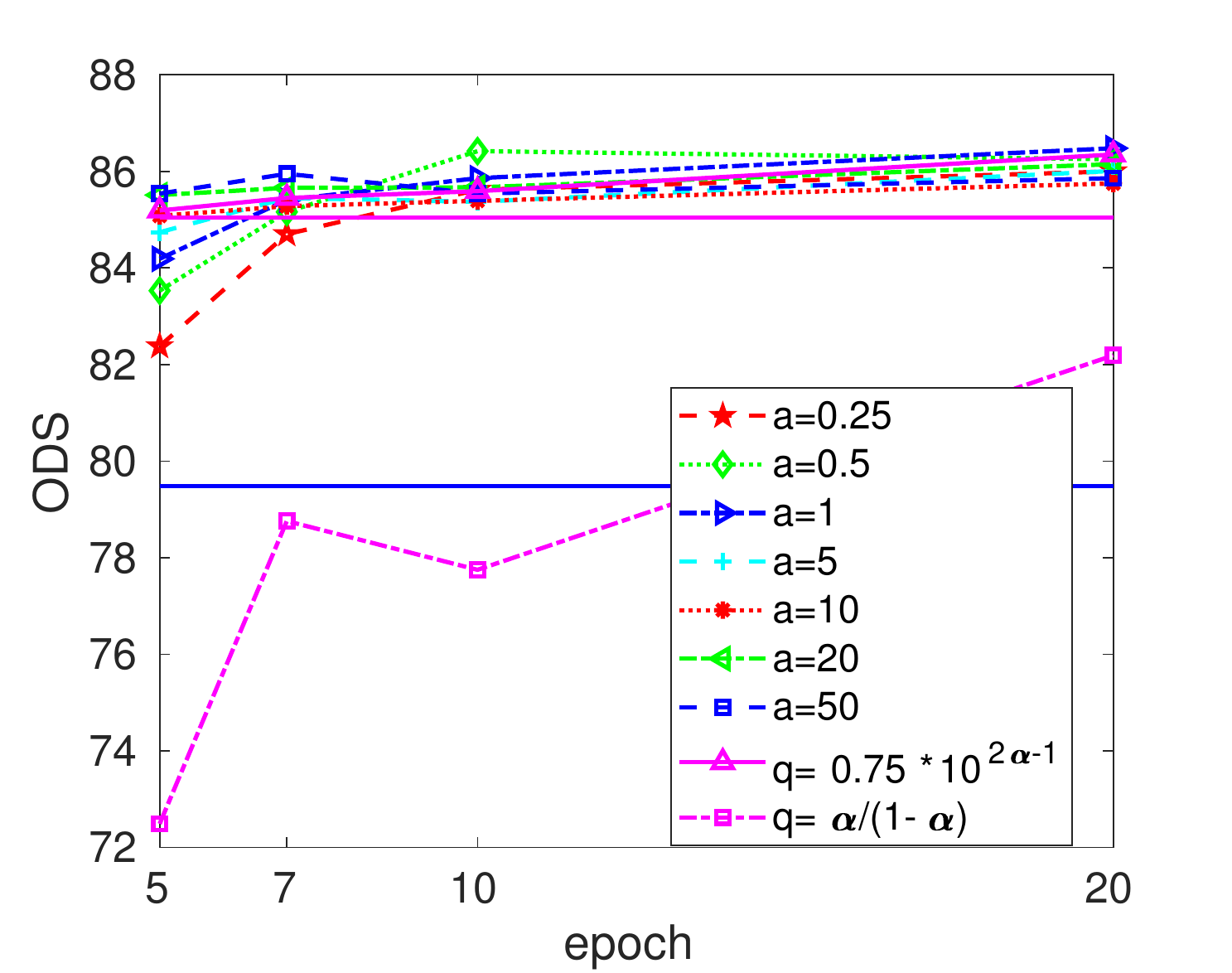}
    }
    \hspace{3cm}
    \subfigure{
    \includegraphics[width=4.3cm]{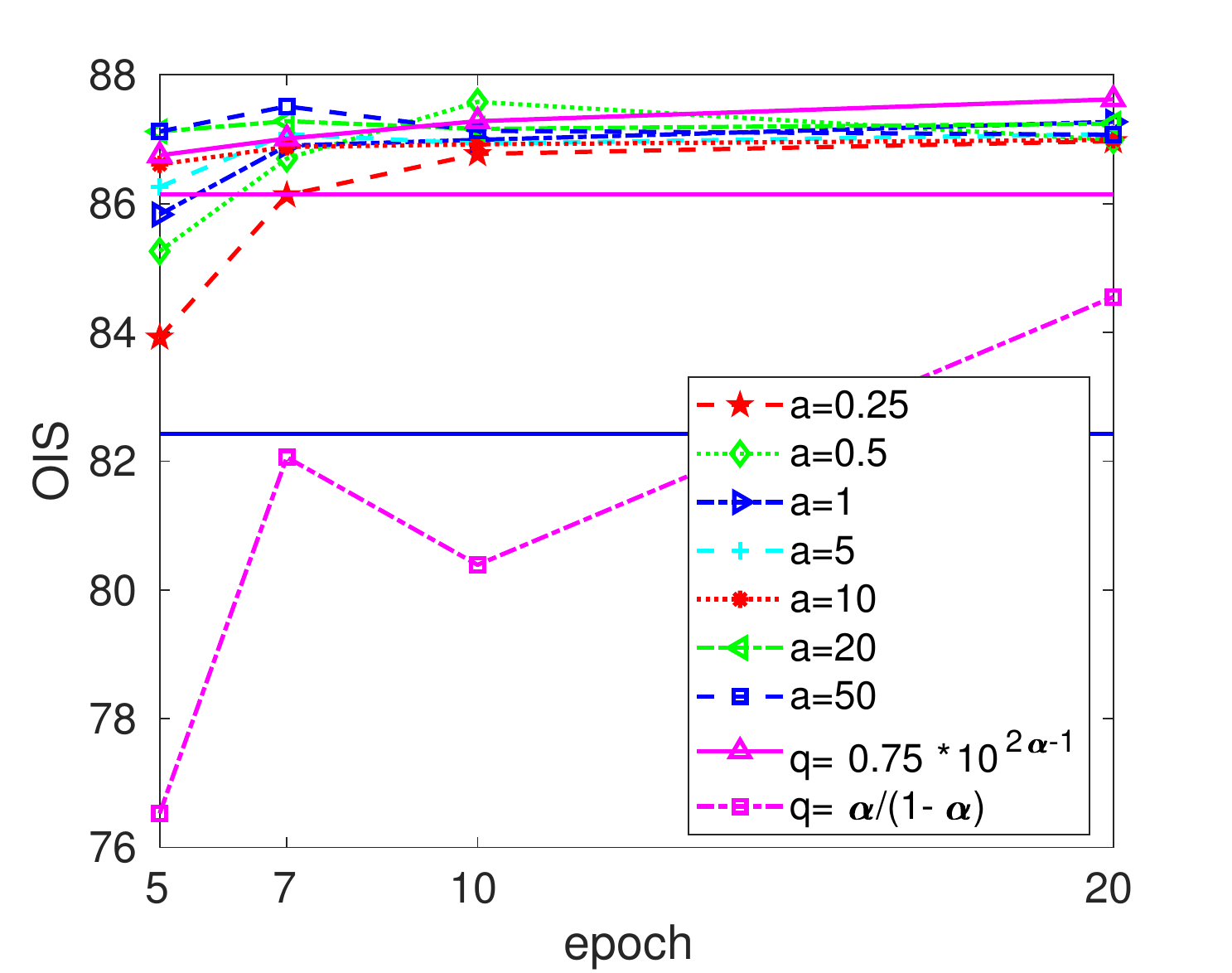}
    }
    \caption{The effect of Jaccard distance. Here, the solid blue and pink lines mean the first and second baselines, respectively. Note that the second baseline is with 70 training epochs. The solid pink line with triangle indicates the exponential WCE ($\beta$=$0.75$). The others correspond to the loss in \eqref{the lotal loss} with fixing $b=1$. }
    \label{The effect of Jaccard distance}
\end{figure*}

\subsubsection{\textbf{The effect of Jaccard-index}}
Now, we investigate the impact of adding a Jaccard-index to the weighted cross-entropy, in correspondence with \eqref{the lotal loss}. Correctly, we reveal it via a trend graph in \reffig{The effect of Jaccard distance}. Besides, we list some specific results in \reftab{fast and accurate results on the first database}.
Then we find that U-Net could converge \emph{earlier} and retain the test accuracy when we add a suitable Jaccard-index item to the weighted cross-entropy.
For example, with a combination of a=20 and b=1, the metrics ODS and OIS obtained by 5 training epochs exceed those of the first and second benchmarks, and concurrently are approximate to those obtained by exponential WCE with 10 training periods.
When fixing training epochs, the test results are \emph{more accurate} by adding it. For example, the F{$_1$} scores are at least 0.3\% higher than those obtained by merely applying exponential WCE.
Besides, some visual results are shown in \reffig{visual comparision on CrackForest}.

{Moreover, we also discuss the reason why our loss functions are valid from the perspective of the Jaccard coefficient and learning rate, respectively. Please see Appendix \ref{Appendix A} for details.}

\begin{figure*}[htbp]
\centering
\includegraphics[width=16.8cm]{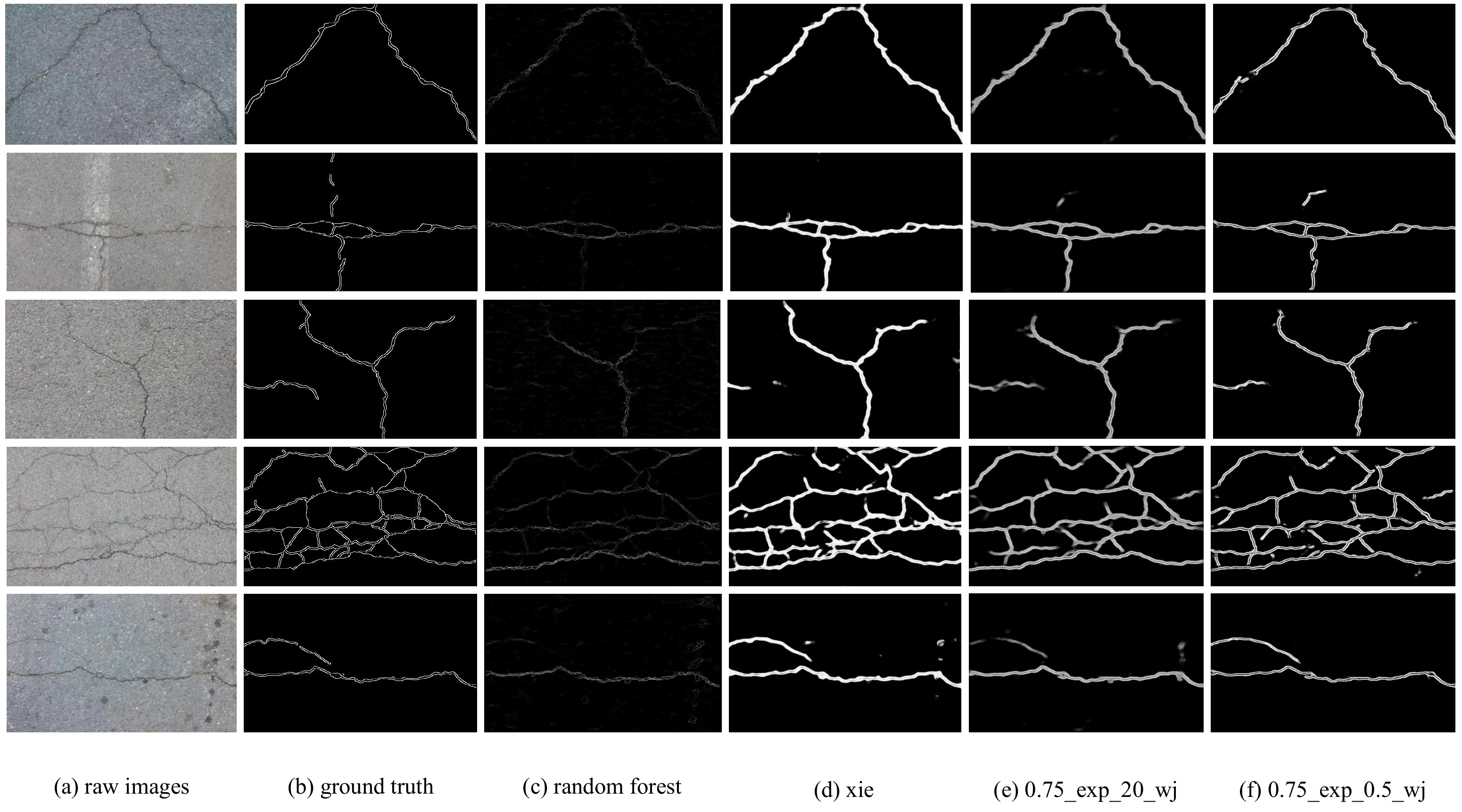}
\caption{Qualitative examples of using different methods on CrackForest. (a) contains the raw images; (b) corresponds to the {ground truth}; (c) and (d) indicate the results obtained by the first and second baselines, respectively; (e$\sim$f) are the ones originated from our loss functions with 5 and 10 training epochs, respectively. Note that the second baseline is with 70 training epochs.}
\label{visual comparision on CrackForest}
\end{figure*}

\begin{table*}[htbp]\footnotesize
\caption{The average results {with 5 runs} obtained by different WCE on {AigleRN}. Here $*$ means applying a Laplacian smoothing item. Note that `$1\_exp\_10\_wj$' indicates applying formula \eqref{the lotal loss} by setting $a=10$ and $b=1$ with exponential WCE ($\beta=1$).}
\centering
\begin{tabular*}{\hsize}{@{}@{\extracolsep{\fill}}c|c|c|c|c|ccc|ccc@{}}
\hline
\multirow{2}{*}{Methods}  & \multirow{2}{*}{$\beta$}  & \multirow{2}{*}{$\gamma$}  & \multirow{2}{*}{epoch} & \multirow{2}{*}{time} & \multicolumn{3}{c|}{ODS}  & \multicolumn{3}{c}{OIS} \\
\cline{6-11}
    & & & &              & P            & R            & F{$_1$}                    & P             & R                   & F{$_1$}  \\
\hline
RSF \cite{shi2016automatic}
& - & -  & -  & -        &85.81	       &88.02	      &86.9	                 &53.95	          &\textbf{95.46}	             &68.94   \\
\hline
\multirow{4}{*}{$wce\_xie*$ \cite{xie2015holistically}}
& - & - & 50  &15min 5s  &85.93 (2.11)  &87.38 (1.68)   &86.62 (0.58)	         &90.35 (1.36)	  &92.03 (0.93)	         &91.18 (1.02)  \\
& &     & 15  &4min 35s  &84.35 (2.25)  &87.77 (3.81)   &85.97 (1.84)	         &78.63 (8.09)	  &93.09 (3.05)	         &85.02 (4.71)  \\
& &     & 7   &2min 11s  &84.79 (3.09)  &86.01 (2.54)   &85.34 (1.52)	         &76.46 (4.74)	  &93.11 (1.71)	         &83.92 (3.32)  \\
& &     & 4   &1min 17s  &83.61 (1.39)  &86.81 (0.85)   &85.17 (0.66)	         &66.55 (2.40)	  &92.80 (0.84)	         &77.50 (1.91)  \\
\hline
\multirow{2}{*}{$ wce\_power* $}  & \multirow{2}{*}{0.5}  & \multirow{2}{*}{1/2}
    &15 &4min 35s   &86.45 (2.05)    &\textbf{88.64} (2.56)	&87.49 (1.14)	&93.38 (1.53)	&89.63 (2.80)	&91.43 (1.39)  \\	
& & &7	&2min 11s   &86.65 (2.30)	&86.83 (2.19)	&86.69 (0.67)	&93.68 (2.05)	&88.35 (2.52)	&90.90 (1.12)	\\
\hline
\multirow{2}{*}{$ wce\_log* $}	& \multirow{2}{*}{1} & \multirow{2}{*}{-}
    &15 &4min 35s    &\textbf{88.05} (1.71)	&86.59 (2.23)	&87.28 (0.42)	&94.03 (1.66)	&86.77 (2.91)	&90.21 (0.96) \\
& & &7	&2min 11s    &87.32 (2.16)	&86.00 (3.59)	&86.58 (1.13)	&\textbf{95.08} (1.89)	&86.47 (4.25)	&90.48 (1.66)  \\
\hline
\multirow{2}{*}{$ wce\_exp* $}	& \multirow{2}{*}{0.75} & \multirow{2}{*}{1}
    &15 &4min 35s   &87.20 (2.12)	&88.31 (3.23)	&87.68 (0.83)	&93.94 (0.92)	&88.72 (3.37)	&91.21 (1.60)   \\
& & &7	&2min 11s   &87.86 (2.72)	&86.27 (4.08)	&86.94 (1.19)	&93.50 (2.56)	&88.49 (4.62)	&90.82 (2.07)   \\
\hline
\multirow{2}{*}{$ wce\_exp $} 	& \multirow{2}{*}{1}	& \multirow{2}{*}{1}	
    &15 &4min 35s	  &87.56 (2.31)	&88.01 (1.99)	&\textbf{87.74} (0.51)	&93.48 (1.12)	&90.78 (2.36)	&\textbf{92.08} (0.99)	\\
& & &7	&2min 11s     &{87.69} (2.07)	&86.70 (1.67)	&87.16 (0.67)	&94.19 (1.23)	&88.75 (2.63)	&91.36 (1.07)	\\
\hline
{ $ 1\_exp\_10\_wj $}	&1  &1
&\textbf{4}	&\textbf{1min 17s}		&86.51 (1.56)	&87.45 (2.32)	&86.94 (0.64)	&92.78 (1.93)	&90.21 (2.30)	&91.45 (1.39)	\\
\hline
\end{tabular*}
\label{The fast and accurate results on AiglRN database}
\end{table*}

\subsection{Performance on AigleRN Dataset}\label{4B}
There are 38 images in AigleRN dataset\cite{2011Automatic}. We utilize 60\% and 40\% of it for training and testing separately.
Since the training images are few and the cracks are sparse in this database,
we first fulfill Gaussian noise with an normal distribution N(0, 0.01) to enlarge training data, and then apply data augmentation as mentioned before.
{Due to insufficient training data, we use the channels [32, 64, 128, 256, 512] in the encoder of U-Net. Besides, AigleRN contains two specifications, i.e., 311 or 991 pixels in width and 462 pixels in height. Note that four down-sampling restricts the input size to be a multiple of 16. Otherwise, the same level feature maps from the encoder and decoder are not equal in size, resulting in the infeasibility of channel concatenation. Thus, we utilize 304$\times$448} as the input of U-Net.

The second baseline is obtained with 50 training epochs (200 steps per epoch). The specific results\footnote{Considering the size difference between the inputs and training images, some inputs may have no crack pixels and then the denominator $1-\alpha$ in \eqref {Power function type} and \eqref{Logarithmic function type} is likely zero for the power and logarithmic function type respectively. Hence, we apply the Laplacian smoothing. Besides, since F{$_1$} score (ODS) on the test set is less than 85\% when utilizing \eqref{the lotal loss} to train U-Net with the hyper-parameter a $ \in (0,1]$, we do not list the related results in \reftab{The fast and accurate results on AiglRN database}.}
are shown in \reftab{The fast and accurate results on AiglRN database}.
Compared with the second benchmark, our weighted cross-entropy can shorten the training time to at least 1/7 of the former while retaining the similar performance. If we combine our loss with the Jaccard-index ($a$=$10$, $b$=$1$), the training time can be further reduced to 1/12 of the original. Besides, we also show some visual results via \reffig{visual comparision on AiglRN} in Appendix \ref{Appendix B}.

\begin{table*}[htbp]\footnotesize
\caption{The average results {with 5 runs} obtained by different WCE on Crack360. Note that `$1\_exp\_10\_wj$' means applying formula \eqref{the lotal loss} by setting $a=10$ and $b=1$ with exponential WCE ($\beta=1$).}
\centering
\begin{tabular*}{\hsize}{@{}@{\extracolsep{\fill}}c|c|c|c|c|ccc|ccc@{}}
\hline
\multirow{2}{*}{Methods}  & \multirow{2}{*}{$\beta$}  & \multirow{2}{*}{$\gamma$}  & \multirow{2}{*}{epoch} & \multirow{2}{*}{time} & \multicolumn{3}{c|}{ODS}  & \multicolumn{3}{c}{OIS} \\
\cline{6-11}
    & & & &              & P            & R            & F{$_1$}                    & P             & R                   & F{$_1$}  \\
\hline
RSF \cite{shi2016automatic}
& - & -  & -  & -        &65.10   &67.01	      &66.04	                 &75.43	          &88.58	             &81.48  \\
\hline
\multirow{4}{*}{$ wce\_xie$ \cite{xie2015holistically}}
& - & -  &70  &87min 35s  &88.71 (3.43)  &90.42 (2.70)   &89.54 (2.86)	      &89.45 (2.06)	  &97.34 (0.46)	       &93.21 (1.13)  \\
&  &   &10  &12min 35s    &84.41 (3.45)  &91.29 (1.63)   &87.70 (2.46)	      &90.04 (1.00)	  &\textbf{97.79} (0.30)	       &93.75 (0.58)   \\
&  &   &5   &6min 20s   &82.65 (2.31)  &87.11 (2.21)   &84.78 (0.77)	      &88.70 (0.96)	  &97.33 (0.46)	       &92.81 (0.64)    \\
&  &   &3   &3min 50s   &69.17 (2.37)  &83.05 (1.90)   &75.46 (1.77)	      &83.44 (1.51)	  &96.83 (0.63)	       &89.63 (1.13)    \\
\hline
\multirow{2}{*}{$ wce\_power $}  & \multirow{2}{*}{0.5}  & \multirow{2}{*}{1/2}
    &10 &12min 35s   &94.30 (0.89)    &{92.47} (0.88)	&93.37 (0.66)	&94.40 (0.54)	&97.41 (0.29)	&\textbf{95.88} (0.25)  \\	
& & &5	&6min 20s   &91.24 (1.89)	&87.14 (1.58)	&89.13 (1.47)	&92.00(1.82)	&96.62 (0.48)	&94.24 (0.96)	\\
\hline
\multirow{2}{*}{$ wce\_log $}	& \multirow{2}{*}{0.9} & \multirow{2}{*}{-}
    &10 &12min 35s    &{93.36} (0.93)	 &92.31 (0.56)	&92.83 (0.59)	&94.20 (1.12)	&97.14 (0.40)	&95.64 (0.53) \\
& & &5	&6min 20s    &90.72 (1.82)	&88.26 (2.88)	&89.46 (2.20)	&{90.76} (1.75)	&96.81 (0.96)	&93.68 (1.06)  \\
\hline
\multirow{2}{*}{$ wce\_exp $}	& \multirow{2}{*}{1} & \multirow{2}{*}{1}
    &10 &12min 35s   &94.06 (0.53)	&\textbf{93.26} (0.70)	&\textbf{93.65} (0.44)	&93.91 (0.62)	&97.29 (0.23)	&95.57 (0.33)   \\
& & &5	&6min 20s   &91.42 (0.53)	&88.42 (1.57)	&89.89 (0.94)	&90.02 (1.06)	&{97.60} (0.53)	&93.65 (0.47)   \\
\hline
\multirow{2}{*}{$ 1\_exp\_10\_wj $}	& \multirow{2}{*}{1} & \multirow{2}{*}{1}
    & {5}	&{6min 20s}		&\textbf{94.47} (0.77)	&90.54 (1.41)	&92.46 (0.84)	&\textbf{96.97} (0.94)	&94.31 (1.04)	&95.61 (0.58)	\\
&& &\textbf{3}	&\textbf{3min 50s}		&89.89 (3.70)	&89.30 (2.92)	&89.57 (2.85)	&93.17 (4.06)	&94.84 (2.07)	&93.93 (1.88)	\\
\hline
\end{tabular*}
\label{The fast and accurate results on Crack360 database}
\end{table*}

\begin{table*}[htbp]\footnotesize
\caption{The average results {with 5 runs} obtained by different WCE on BJN260. Note that `$power\_2\_wj$' indicates applying formula \eqref{the lotal loss} by setting $a=2$ and $b=1$ with power WCE ($\beta=0.5, \gamma=1/2$).}
\centering
\begin{tabular*}{\hsize}{@{}@{\extracolsep{\fill}}c|c|c|c|c|ccc|ccc@{}}
\hline
\multirow{2}{*}{Methods}  & \multirow{2}{*}{$\beta$}  & \multirow{2}{*}{$\gamma$}  & \multirow{2}{*}{epoch} & \multirow{2}{*}{time} & \multicolumn{3}{c|}{ODS}  & \multicolumn{3}{c}{OIS} \\
\cline{6-11}
    & & & &              & P            & R            & F{$_1$}                    & P             & R                   & F{$_1$}  \\
\hline
RSF \cite{shi2016automatic}
& - & -  & -  & -        &45.08   &45.93	      &45.50	                 &38.49	          &\textbf{51.77}	             &44.15   \\
\hline
\multirow{4}{*}{$ wce\_xie$ \cite{xie2015holistically}}
& - & -  &30  &24min  &65.82 (1.89)  &40.25 (0.60)   &49.94 (0.36)	      &59.84 (2.38)	  &41.93 (0.50)	       &49.28 (0.66)  \\
&  &   &15  &12min    &66.22 (1.01)  &38.90 (0.52)   &49.01 (0.60)	      &51.39 (0.98)	  &{41.10} (0.68)	       &45.67 (0.73)   \\
&  &   &7   &5min 36s   &58.49 (2.18)  &35.56 (1.46)   &44.23 (1.70)	      &46.28 (0.76)	  &37.02 (2.33)	       &41.10 (1.30)    \\
&  &   &5   &4min   &54.15 (2.56)  &33.58 (2.21)   &41.44 (2.27)	      &44.67 (1.33)	  &34.48 (2.64)	       &38.90 (2.20)    \\
\hline
\multirow{2}{*}{$ wce\_power $}  & \multirow{2}{*}{0.5}  & \multirow{2}{*}{1/2}
    &15 &12min   &67.97 (1.56)    &{40.17} (0.90)	&50.48 (0.30)	&64.27 (1.64)	&41.77 (0.77)	&{50.62} (0.34)  \\	
& & &7	&5min 36s   &63.77 (1.76)	&\textbf{41.19} (0.64)	&50.05 (0.90)	&61.68(1.19)	&42.80 (0.79)	&50.54 (0.88)	\\

\hline
\multirow{2}{*}{$ wce\_log $}	& \multirow{2}{*}{0.9} & \multirow{2}{*}{-}
    &15 &12min    &\textbf{68.31} (0.93)	 &40.28 (0.51)	&50.67 (0.32)	&64.27 (1.38)	&42.16 (0.57)	&\textbf{50.91} (0.33) \\
& & &7	&5min 36s    &64.83 (1.69)	&40.73 (0.88)	&50.02 (0.76)	&{62.42} (0.81)	&42.64 (0.90)	&50.66 (0.81)  \\
\hline
\multirow{2}{*}{$ wce\_exp $}	& \multirow{2}{*}{0.75} & \multirow{2}{*}{1}
    &15 &12min  &67.71 (0.94)	&{40.72} (0.57)	&\textbf{50.85} (0.42)	&62.27 (1.01)	&42.52 (0.60)	&50.52 (0.42)   \\
& & &7	&5min 36s   &62.80 (1.37)	&40.47 (0.62)	&49.22 (0.81)	&60.50 (1.38)	&{41.72} (1.07)	&49.38 (1.07)   \\

\hline
\multirow{1}{*}{$ power\_2\_wj $}	& \multirow{1}{*}{0.5} & \multirow{1}{*}{1/2}
    & \textbf{5}	&\textbf{4min}		&{63.70} (1.95)	&40.38 (1.19)	&49.40 (0.72)	&\textbf{65.01} (1.33)	&41.14 (1.17)	&50.37 (0.71)	\\
\hline
\end{tabular*}
\label{The fast and accurate results on BJN260 database}
\end{table*}

\subsection{Performance on Crack360 Dataset}\label{4C}
Here we utilize the road crack dataset CrackTree260 \cite{zou2018deepcrack} for training U-Net, and CRKWH100 \cite{zou2018deepcrack} for testing. In detail, the training set contains 260 images where each of them has 800$\times$600 pixels. The test set has 100 {pictures} where every one has $512\times512$ pixels. During training, every image is cropped into $512\times512$ pixels. For convenience, we unify these two datasets into one database, Crack360.

To get the second baseline, we utilize the weighted method in \eqref{q_xie} to train U-Net for 70 epochs. The specific evaluations are shown in \reftab{The fast and accurate results on Crack360 database} and visual results via \reffig{visual comparision on Crack360} in Appendix \ref{Appendix B}. Compared with {the second benchmark}, our weighted cross-entropy can shorten the training time to 1/14 of the former while retaining the similar performance. Besides, the training time is further reduced to at least 1/23 of the former if one combines our loss with the Jaccard distance ($a$=$1$, $b$=$1$).

\subsection{{Performance on BJN260 Dataset}\label{4D}}
{BJN260 is a pavement crack database in Beijing's night scenes. This database contains 260 images and corresponding pixel-level annotations. These crack images are captured by a mobile phone, HUAWEI Honor 6X.  They share the same size of 480 and 320 pixels in width and height, respectively. We randomly choose 200 images for training, and the rest for testing. Different from the daytime scenes, the night ones are more complex and changeable. Thus, these cracks are more difficult to distinguish.}

{To obtain the second baseline, we utilize the weighted method in \eqref{q_xie} to train U-Net for 30 epochs. Please see the specific evaluations are shown in \reftab{The fast and accurate results on BJN260 database}. Compared with the second baseline, our weighted cross-entropy can shorten the training time to 1/4 of the former while retaining the approximate performance. Meanwhile, the training time is further reduced to at least 1/6 of the former if one combines our loss with the Jaccard distance ($a$=$2$, $b$=$1$). Besides, we also show some visual results via \reffig{visual comparision on BJN260} in Appendix \ref{Appendix B}.}

\begin{table*}[htb]\footnotesize
    \caption{Comparison with other models on CrackForest. Here, `$\dag$' refers to utilizing our exponential WCE. Besides, {$\sharp$ para} represents the model parameters. `GFLOPS' indicates giga floating-point operations per second. `TT' means the training time. `FPS' represents the average frames per second during testing.}
    \centering
    \begin{tabular*}{\hsize}{@{}@{\extracolsep{\fill}}c@{}|c@{\hspace{0.75cm}}|c@{\hspace{0.75cm}}|c@{\hspace{0.75cm}}|c@{\hspace{0.75cm}}|c@{\hspace{0.75cm}}|c@{\hspace{0.75cm}}|ccc@{\hspace{0.75cm}}}
    \hline
    \multirow{1}{*}{Methods}    &{$\beta$}      &{ODS}         &{OIS}                  &{$\sharp$ para}   &{GFLOPS}  & TT   & FPS         \\
    \hline
    {U-Net \cite{ronneberger2015u}}	         &-     &\textbf{85.04}  	       &\textbf{86.14}         &31.03M 	         & 128.18    &\textbf{54min 54s}          &\textbf{32.27}      	\\
    {U-CliqueNet {\cite{Li2020Automatic}}}   &-       &84.95  	       &86.06         &\textbf{487.77K} 	     &\textbf{38.52}     & 2h 29min          &25.86	    \\
    \hline
    {U-Net{$^\dag$}}	      &0.75             &{86.35}  	       &\textbf{87.62}         &31.03M 	         & 128.18    &{\textcolor{green}{15min 44s}}          &{32.27}      	\\
    {U-CliqueNet{$^\dag$}}    &1              &\textbf{86.41}  	      &87.50         &{487.77K} 	     & {38.52}     & 1h 25min        &25.86	    \\
    \hline
    \end{tabular*}
    \label{Comparison on CrackForest}
\end{table*}

{\begin{table*}[htb]\footnotesize
    \caption{Comparison with other models on {AigleRN}. Here, `$\dag$' refers to applying our exponential WCE.}
    \centering
    \begin{tabular*}{\hsize}{@{}@{\extracolsep{\fill}}c@{}|c@{\hspace{0.85cm}}|c@{\hspace{0.75cm}}|c@{\hspace{0.75cm}}|c@{\hspace{0.75cm}}|c@{\hspace{0.75cm}}|c@{\hspace{0.75cm}}|ccc@{\hspace{0.75cm}}}
    \hline
    \multirow{1}{*}{Methods}    &{$\beta$}   & ODS        & OIS      &{$\sharp$ para}   &{GFLOPS}  & TT   & FPS         \\
    \hline
    {U-Net \cite{ronneberger2015u}}	         &-      &\textbf{86.62}       &\textbf{91.18}      &7.76M 	       	&28.44       &\textbf{15min 5s}    &\textbf{11.90}   \\
    {U-CliqueNet {\cite{Li2020Automatic}}}   &-      &83.31  	&90.68        &\textbf{487.77K} 	       	&\textbf{34.06}       &20min 40s 	&9.97   \\
    \hline
    {U-Net{$^\dag$}}                       &1       &\textbf{87.74}  	&\textbf{92.08}       &7.76M 	       	&28.44       &\textcolor{green}{\textbf{4min 35s}}    &11.90  \\
    {U-CliqueNet{$^\dag$}}                 &1       &83.34      &90.82       &487.77K 	       	&34.06       &\textcolor{green}{10min 20s}     	&9.97      \\
    \hline
    \end{tabular*}
    \label{Comparison on AiglRN}
\end{table*}}

{\begin{table*}[htb]\footnotesize
    \caption{Comparison with other models on Crack360. Here, `$\dag$' refers to utilizing our exponential WCE.}
    \centering
    \begin{tabular*}{\hsize}{@{}@{\extracolsep{\fill}}c@{}|c@{\hspace{0.75cm}}|c@{\hspace{0.75cm}}|c@{\hspace{0.75cm}}|c@{\hspace{0.75cm}}|c@{\hspace{0.75cm}}|c@{\hspace{0.75cm}}|ccc@{\hspace{0.75cm}}}
    \hline
    \multirow{1}{*}{Methods}  &{$\beta$}   & ODS        & OIS      & {$\sharp$ para}   &{GFLOPS}  &TT     & FPS     \\
    \hline
    {FCN8s \cite{long2015fully}}        &-   &91.23    &93.66     &134.27M      &189.5        &2h 54min    & 13.92  \\
    {HED \cite{xie2015holistically}}        &-   &92.48    & 95.22    &14.72M       &80.46        &1h 42min   &\textbf{31.80}     \\
    {RCF \cite{liu2017richer}}              &-   &92.71    &94.93     &14.8M        &102.67      &3h 55min    &23.87     \\
    {SegNet \cite{badrinarayanan2017segnet}}&-   &94.36    &96.86     &29.44M       &160.11      &2h 32min    &26.21      \\
    {{DeepCrack \cite{zou2018deepcrack}}}   &-   &\textbf{94.59}    &\textbf{96.89}     & 29.48M     & 170.1    &3h 13min   &19.73      \\
    \hline
    {Deeplab v3+ \cite{2018Encoder}}       &-   &90.52    & 94.90    &40.35M      &101.16        & 2h 30min     & 20.29    \\  
    {CrackSeg \cite{song2020automated}}     &-   &94.31    &96.88     &53.87M      &197.47      & 3h 48min    & 11.47     \\
    \hline
    {U-Net \cite{ronneberger2015u}} 	    &-    &89.54  & 93.21  &31.03M   &218.46   &\textbf{1h 28min}   & 23.35       \\
    {U-CliqueNet {\cite{Li2020Automatic}}}  &-    &85.47  	&93.18          &\textbf{487.77K}      &\textbf{65.57}       &3h 58min	   & 17.35   \\
    \hline
    {FCN8s{$^\dag$}}            &1          &\textbf{\textcolor{green}{95.01}}       &\textcolor{green}{96.56}     &134.27M      &189.5        &\textcolor{green}{1h 27min}    & 13.92  \\
     {HED{$^\dag$}}             &1          &92.58       &95.91     &14.72M       &80.46       &\textcolor{green}{34min}       &{31.80}     \\
     {RCF{$^\dag$}}             &0.75       &93.60       &95.24     &14.8M        &102.67      &\textcolor{green}{1h 57min}     &23.87     \\
    {SegNet{$^\dag$}}           &0.5        &94.94       &96.67     &29.44M       &160.11      &\textcolor{green}{1h 4min}      &26.21      \\
    {DeepCrack{$^\dag$}}        &0.75       &95.00       &96.41     &29.48M       &170.1      & 2h 9min      &19.73      \\
    {Deeplab v3+{$^\dag$}}      &1          &90.34       &96.10     &40.35M       &101.16        & 2h 30min     & 20.29    \\  
    {CrackSeg{$^\dag$}}         &0.75       &94.96       &\textbf{97.49}     &53.87M       &197.47      &\textcolor{green}{1h 54min}     &11.47     \\
    {U-Net{$^\dag$}} 	        &1          &\textcolor{green}{93.65}       &\textcolor{green}{95.57}     &31.03M       &218.46   &\textcolor{green}{\textbf{13min}}     &23.35       \\
    {U-CliqueNet{$^\dag$}}      &1          &\textcolor{green}{93.26}  	 &\textcolor{green}{95.86}     &{487.77K}    &{65.57}       &\textcolor{green}{1h 8min}	   &17.35   \\
    \hline
    \end{tabular*}
    \label{Comparison on Crack360}
\end{table*}}

{\begin{table*}[htb]\footnotesize
    \caption{Comparison with other models on BJN260. Here, `$\dag$' refers to applying our exponential WCE.}
    \centering
    \begin{tabular*}{\hsize}{@{}@{\extracolsep{\fill}}c@{}|c@{\hspace{0.75cm}}|c@{\hspace{0.75cm}}|c@{\hspace{0.75cm}}|c@{\hspace{0.75cm}}|c@{\hspace{0.75cm}}|c@{\hspace{0.75cm}}|ccc@{\hspace{0.75cm}}}
    \hline
    \multirow{1}{*}{Methods}                    &{$\beta$}  & ODS        & OIS   & {$\sharp$ para}    &{GFLOPS}   & TT   & FPS         \\
    \hline
    {FCN8s \cite{long2015fully}}            &-    &50.11         &51.02   &134.27M   &129.53   &3h 9min    &20.89    \\
    {HED \cite{xie2015holistically}}            &-    &50.03        &52.23   &14.72M   &47.32      &1h 3min    &\textbf{41.45}    \\
    {RCF \cite{liu2017richer}}                  &-    &\textbf{52.52}    &\textbf{54.74}  &14.8M   &60.16      & 5h 32min      &32.43   \\
    {SegNet \cite{badrinarayanan2017segnet}}    &-    &44.91    &46.74   & 29.44M   &93.99      &3h 25min       &35.50     \\
    {{DeepCrack \cite{zou2018deepcrack}}}       &-     &48.17   &50.22  & 29.48M   &99.85      & 2h 50min      & 27.64      \\
    \hline
    {Deeplab v3+ \cite{2018Encoder}}            &-     &44.63    &46.12   &40.35M    &59.76      &1h 6min      &26.88     \\
    {CrackSeg \cite{song2020automated}}         &-     &47.34    &48.10  &53.87M  &116.67      &1h      &15.96     \\
    \hline
    {U-Net \cite{ronneberger2015u}}             &-     &49.94  	&49.28   &31.03M  	&128.18   &\textbf{24min}  &32.27 	  \\
    {U-CliqueNet {\cite{Li2020Automatic}}}      &-     &48.97  	&48.88   &\textbf{487.77K}  	&\textbf{38.52}    &52min   &25.86	  \\
    \hline
    {FCN8s{$^\dag$}}            &0.75   &\textcolor{green}{52.35}         &\textcolor{green}{53.14}   &134.27M   &129.53   &2h 8min    &20.89    \\
    {RCF{$^\dag$}}              &1      &\textcolor{green}{\textbf{54.88}}    &\textcolor{green}{\textbf{57.50}}  &14.8M   &60.16      & 4h 15min      &32.43   \\
    {Deeplab v3+{$^\dag$}}      &0.75   &46.46          &46.26   &40.35M    &59.76      &1h      &26.88     \\
    {CrackSeg{$^\dag$}}         &0.75   &47.33          &48.46  &53.87M  &116.67      &40min      &15.96     \\
    {U-Net{$^\dag$}} 	        &0.75   &50.85          &50.52  &31.03M  &128.18   &\textcolor{green}{\textbf{12min}}      &32.27       \\
    {U-CliqueNet{$^\dag$}}      &0.75   &50.67  	    &49.82   &{487.77K}  	&{38.52}    &31min  &25.86	  \\
    \hline
    \end{tabular*}
    \label{Comparison on BJN260}
\end{table*}}

\section{Conclusion and Discussion}\label{part5}
To cope with the imbalanced data in crack detection, we introduce three adaptive weighted cross-entropy losses, based on the cost-sensitive learning mechanism. Experiments on four benchmark databases demonstrate that our methods could accelerate training significantly while retaining the performance of crack detection.

Through \ref{4A}, \ref{4B}, \ref{4C} and \ref{4D}, compared with the power and logarithm types, the exponential weighted method has three advantages:
\begin{itemize}
\item In terms of training, it has broader application scenarios. For example, unlike the latter two, it has no need to consider a Laplacian smoothing in the weighted function when confronting with samples without cracks.
\item Concerning F{$_1$} score based on optimal dataset scale, the exponential type is always better than the power type and the logarithm type through the comparisons in \reftab{fast and accurate results on the first database}, \reftab{The fast and accurate results on AiglRN database}, \reftab{The fast and accurate results on Crack360 database} and \reftab{The fast and accurate results on BJN260 database}.
\item More importantly, it has a clear upper bound (namely $10$, {according to \eqref{Exponential function type}}), {which may have certain reference value for dealing with category imbalance problems.}
\end{itemize}

{Meanwhile, we compare our approach with other models on the above databases, shown in \reftab{Comparison on CrackForest}, \reftab{Comparison on AiglRN}, \reftab{Comparison on Crack360}, and \reftab{Comparison on BJN260}, respectively. As there are too few training images in the first two datasets (only 72 and 23 images, respectively), we merely apply U-Net \cite{ronneberger2015u} and U-CliqueNet {\cite{Li2020Automatic}} and do not utilize the models based on VGG16 \cite{simonyan2015very} and ResNet50 \cite{he2016res}. Note that there are 4 layers in the clique blocks of U-CliqueNet. Besides, FCN8s \cite{yang2018automatic}, HED \cite{xie2015holistically}, RCF \cite{liu2017richer}, SegNet \cite{badrinarayanan2017segnet}, and {DeepCrack \cite{zou2018deepcrack}} are based on VGG16. Deeplab v3+ \cite{2018Encoder} and CrackSeg \cite{song2020automated} are based on ResNet50.}

{To illustrate the performance of different models, apart from test accuracy, we also display the complexities, training time and test speeds of different models. The complexity of a model includes space complexity and time complexity, which are measured by the model parameters and floating-point operations per second. The test speed of a model is evaluated by the average frames per second during testing.}

{By virtue of \reftab{Comparison on CrackForest}, \reftab{Comparison on AiglRN}, \reftab{Comparison on Crack360}, and \reftab{Comparison on BJN260}{\footnote{{Note that in terms of the same model combined with different loss functions, we apply green fonts to emphasize the relevant results when its ODS or OIS increases by no less than 2\% or its training time is reduced by at least half.}}, we find that compared with the lightweight network U-CliqueNet on the above four databases, U-Net has higher test accuracy (in terms of ODS and OIS), less training time, and faster test speed, despite higher model complexity.}}

{On Crack360, compared with the models based on VGG16, U-Net obtains lower test accuracy (in terms of ODS and OIS) and almost higher model complexity, but it requires less training time.  When combined with our proposed loss, U-Net significantly reduces its training time and narrows considerably the gap of test accuracy between it and the state-of-the-art model DeepCrack. One can come to a similar conclusion when comparing U-Net with Deeplab v3+ and CrackSeg, respectively.  Meanwhile, U-Net has fewer model parameters, less training time, and higher test speed.}

{Different from Crack360, BJN260 includes multi-scale cracks, especially trivial cracks, as shown in \reffig{visual comparision on BJN260}. Compared with RCF on BJN260, U-Net and U-CliqueNet have a disadvantage in the metrics ODS and OIS, even through combined with our proposed loss. The main reason is that they make no full use of multi-scale information. Besides, Deeplab v3+ and CrackSeg obtain low metrics ODS and OIS, since they apply no feature information before the first and second down-sampling. SegNet and DeepCrack may lose some information of trivial cracks, as they perform five down-sampling. In contrast, RCF conducts three down-sampling. Although FCN does five down-sampling, it gets higher ODS and OIS than U-Net. This may be because the last three convolutional layers that replace the fully connected layers play an important role in the architecture. Besides, FCN has the highest model complexity in \reftab{Comparison on BJN260}.}

{Since focusing on the loss function dealing with category imbalance in this paper, we apply the exponential WCE to other models to further discuss. Through \reftab{Comparison on CrackForest}, \reftab{Comparison on AiglRN}, \reftab{Comparison on Crack360} and \reftab{Comparison on BJN260}, we can find that most of the models improve the test accuracy to some extent when combined with the exponential WCE instead of the original loss. Meanwhile, their training time is also shortened to varying degrees. Especially on Crack360, U-CliqueNet improves about 7.8\% ODS and 2.7\% OIS respectively while the training time nearly reduces to 1/4 of the original. However, there are also several cases of failure when some models are combined with our loss function, such as SegNet and DeepCrack on Crack360. Besides the distribution of data and the design of the networks as mentioned above, the failure may also be related to our strategy on the loss function. For example, we usually apply quartiles to fine-tune the hyper-parameter $\beta$ in the exponential WCE \eqref{Exponential function type}. Obviously, this strategy is too simple. One can apply annealing strategies \cite{TSAI2020106068} or Bayesian optimization \cite{wu2019practical} to obtain better results.}

\section*{Acknowledgment}
{The authors would like to thank all the editors and anonymous reviewers for their careful reading and insightful remarks.}


\bibliographystyle{IEEEtran}
\bibliography{IEEEabrv,crack_1}

\begin{IEEEbiography}[{\includegraphics[width=1in,height=1.25in,clip,keepaspectratio]{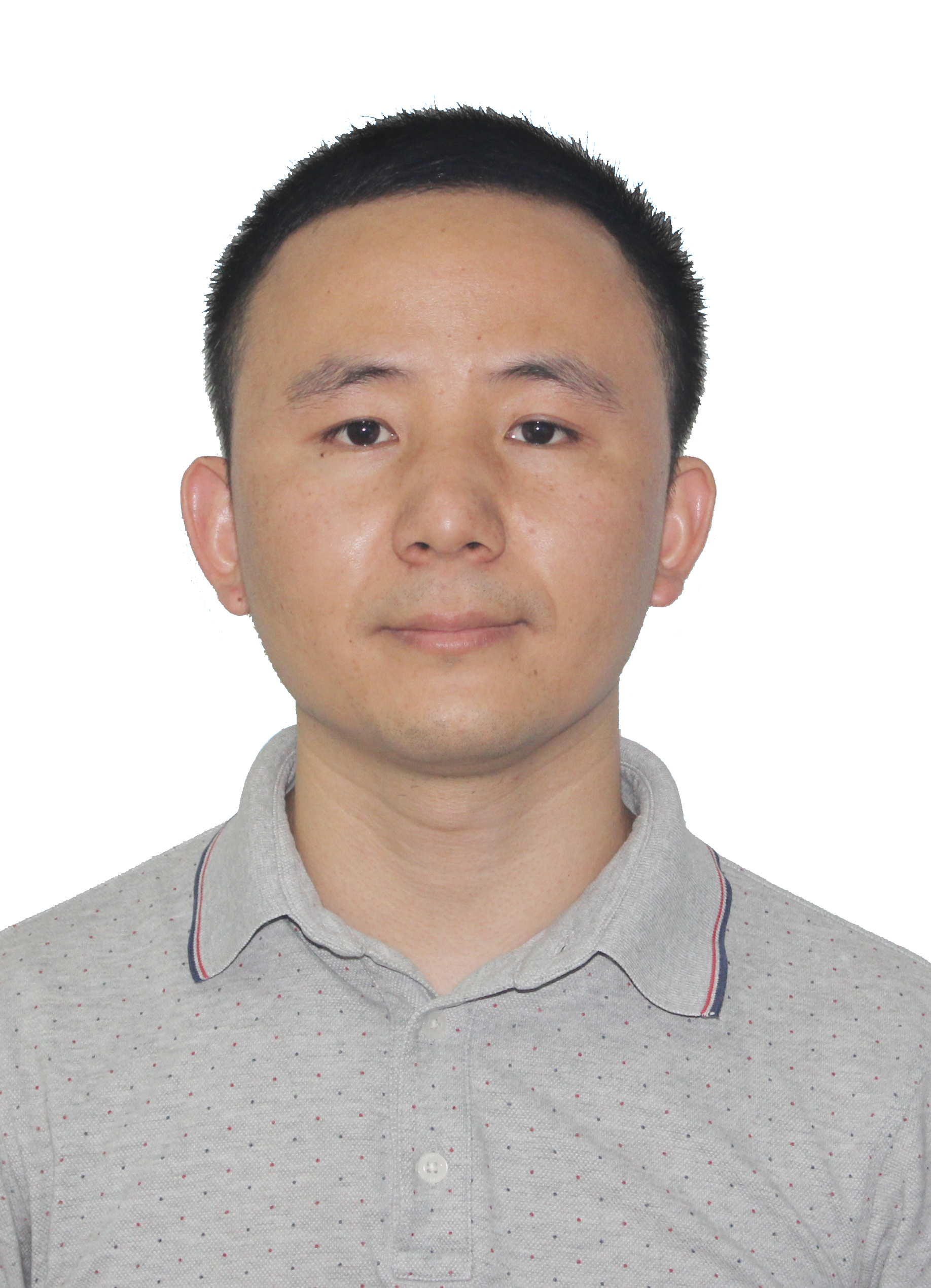}}]{Kai Li}
is working toward the Ph.D. degree with University of Chinese Academy of Sciences, Beijing, China, with a focus on computer vision and machine learning.

His principal research interests include edge detection, image semantic segmentation, and object detection.
\end{IEEEbiography}
\begin{IEEEbiography}[{\includegraphics[width=1in,height=1.25in,clip,keepaspectratio]{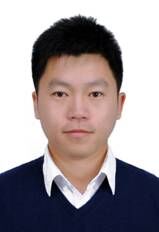}}]{Bo Wang}
received the master’s degree from Beijing Institute of Technology, Beijing, China, in 2010, and the Ph.D. degree from University of Chinese Academy of Sciences, Beijing, in 2014. He was also a visiting scholar in Department of Computer Science and Engineering, Texas A\&M University, in 2019.
He is currently an associate professor with the School of Information and Technology and Management, University of International Business and Economics, Beijing, China.

His principal research interests include statistical machine learning, optimization based data mining, and computer vision.

\end{IEEEbiography}
\begin{IEEEbiography}[{\includegraphics[width=1in,height=1.25in,clip,keepaspectratio]{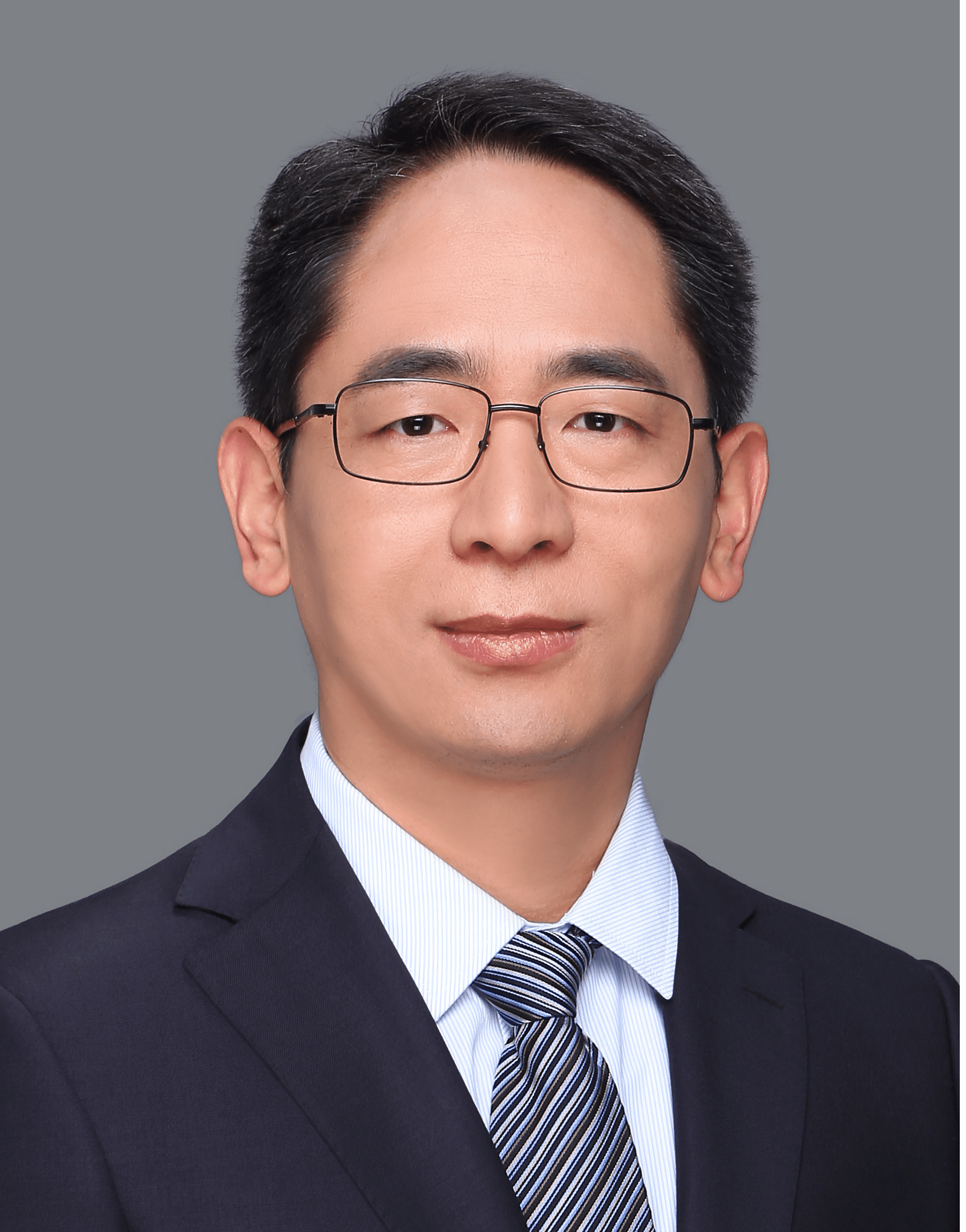}}]{Yingjie Tian} is a Professor with the Research Center on Fictitious Economy and Data Science, Chinese Academy of Sciences. He has published four books on SVMs.

His research interests include support vector machines, optimization theory and its applications, data mining, intelligent knowledge management, and risk management.
\end{IEEEbiography}
\begin{IEEEbiography}[{\includegraphics[width=1in,height=1.25in,clip,keepaspectratio]{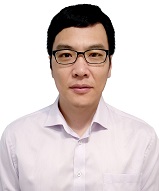}}]{Zhiquan Qi}
is an associate Professor with the Research Center on Fictitious Economy and Data Science, Chinese Academy of Sciences, Beijing, China.

His research interests include image semantic segmentation, super resolution, object detection, object tracking, change detecting, and statistical machine learning.
\end{IEEEbiography}

\newpage
\appendix

\subsection{{The validity of the loss function}}\label{Appendix A}
In this section, we discuss why our losses are valid from the perspective of the Jaccard coefficient and learning rate, respectively.
\subsubsection{Perspective based on the Jaccard coefficient}\label{Appendix A1}
Due to the influence of different weighted methods on the training process, the value range of the loss {functions} somewhat varies, which thus is not convenient for the comparison.
Fortunately, the numerical value of the Jaccard coefficient (or Jaccard-index) is fixed in [0,1]. Meanwhile, the Jaccard coefficient is often used to measure the similarity between two images. Thus, we apply it to illustrate the effect of different loss functions during training.

To this end, we conduct some experiments on CrackForest, which are based on U-Net combined with different loss functions. Through \reffig{The validity of the loss function}, one may find that the metrics obtained from our proposed exponential WCE are much larger than that from the second benchmark when fixing the training epochs. Meanwhile, as the training epochs increase gradually, the metrics obtained from our WCE are improved more steadily than those from the second benchmark. Besides, combing a suitable Jaccard-index item with our proposed WCE can further help to train the model fast and steadily.

\begin{figure}[htbp]
\centering
\subfigure{
\includegraphics[width=5.9cm]{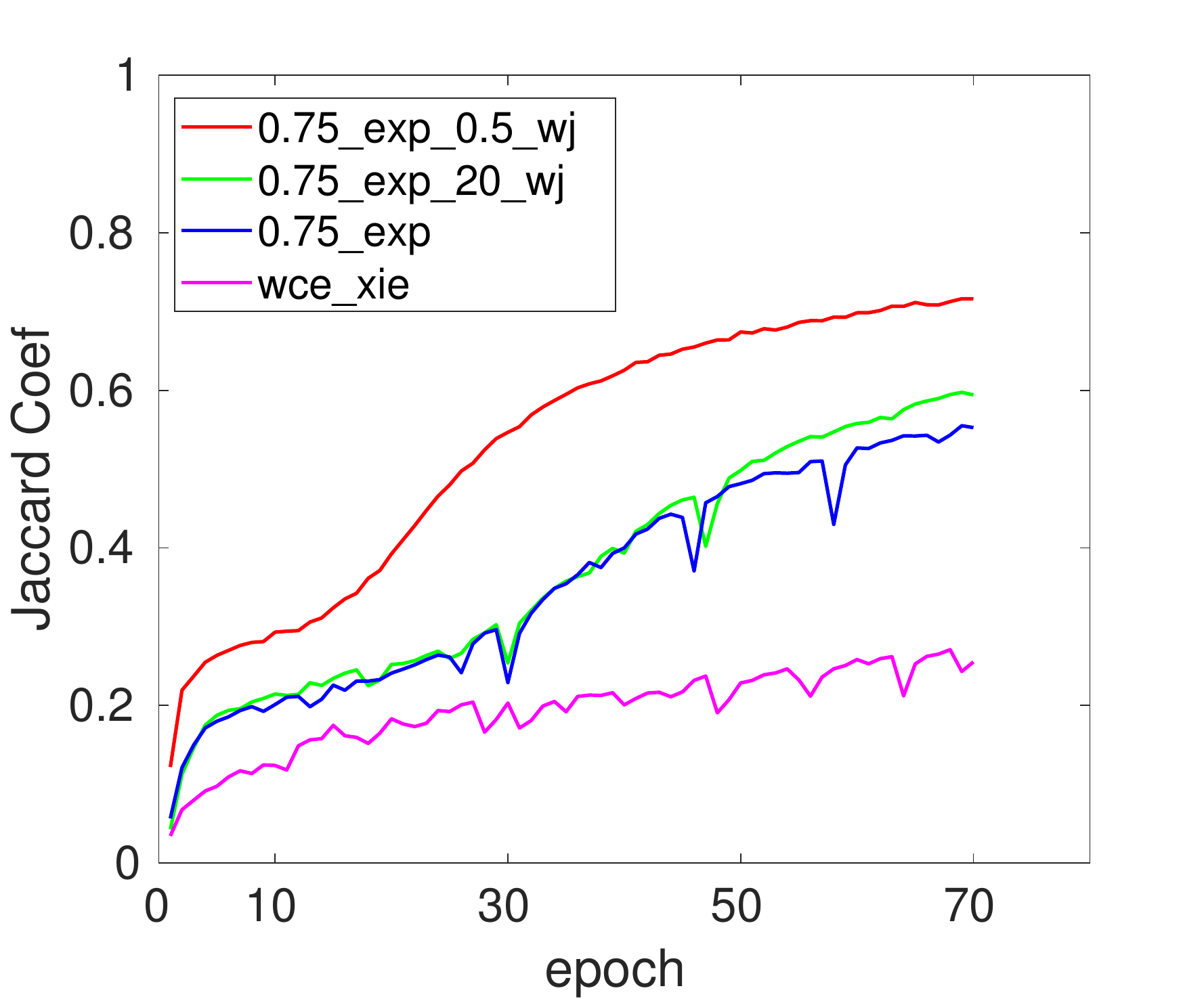}
}
\centering
\caption{The validity of the proposed loss functions. The solid pink line `wce\_xie' refers to the results of the second baseline, i.e., $q$=$\alpha$/(1-$\alpha)$. The solid blue line `0.75\_exp' corresponds to those of the exponentially weighted cross-entropy with $\beta$=$0.75$. The solid green and red lines, respectively corresponding to `0.75\_exp\_20\_wj' and `0.75\_exp\_0.5\_wj', refer to those of the combination of the exponentially weighted cross-entropy ($\beta$=$0.75$) and Jaccard distance, where the combination coefficients are individually chosen as $a$=20, $b$=$1$ and $a$=$0.5$, $b$=$1$.}
\label{The validity of the loss function}
\end{figure}

\subsubsection{{Perspective based on learning rate}}\label{Appendix A2}
{To explore the adaptive nature of our proposed WCE, we start from the perspective of learning rate. According to formula \eqref{eq1}, the loss of a single pixel is that:
\begin{equation}\label{single_pixel_wce}
    l(y_i, p_i) = -\big[q(\alpha) \, y_i \, log \, p_i + (1-y_i) \, log(1-p_i)\big].
\end{equation}
Then, one could obtain the partial derivative of loss $l$ w.r.t the network parameter set $\mathbb{W}$:
\begin{equation}\label{pde1}
  \frac{\partial l}{\partial \mathbb{W}} = - [ q(\alpha) \cdot \frac{y_i}{p_i} - \frac{ 1 - y_i}{1 - p_i}] \cdot \frac{\mathrm{d} {p_{i}}}{\mathrm{d} z_i} \cdot \frac{\partial z_i}{\partial \mathbb{W}}.
\end{equation}
For a binary classification problem, the posterior probability $p_i$ is the sigmoid function of the network output $z_i$, i.e.,
\begin{equation}
   p_i = P(x_i,\mathbb{W}) = exp(z_i)/(1+exp(z_i)),
\end{equation}
where $x_i$ is the i-th pixel value corresponding to $y_i$. Then, one could obtain the derivative of $p_i$ w.r.t $z_i$,
\begin{equation}\label{pde}
  \frac{\mathrm{d} {p_{i}}}{\mathrm{d} z_i} = {p_{i}} \cdot (1-p_{i}).
\end{equation}
Bringing formula \eqref{pde} into \eqref{pde1}, one can obtain
\begin{equation}\label{pde2}
  \frac{\partial l}{\partial \mathbb{W}} =  \frac{\partial z_i}{\partial \mathbb{W}} \, \big[(1 - y_i) \, {p_{i}} -  q(\alpha)\, y_i \,{(1 - p_{i})}\big].
\end{equation}
}

{For ease of statement, the above formula \eqref{pde2} can be written in the following form:
\begin{equation}\label{pde3}
   \frac{\partial l}{\partial \mathbb{W}} =
  \begin{dcases}
     {p_{i}} \,\, \begin{normalsize} {\frac{\partial {z_i}}{\partial {\mathbb{W}}}} \end{normalsize}, & if \,\, y_i=0. \\
     - q(\alpha) \,\, (1 - p_{i}) \,\, \begin{normalsize} {\frac{\partial {z_i}}{\partial {\mathbb{W}}}} \end{normalsize}, & if \,\, y_i=1.
   \end{dcases}
\end{equation}
Note that the first-order updating rule of parameters is that
\begin{equation}\label{update rule}
\mathbb{W} \leftarrow \mathbb{W} - \theta \cdot \frac{\partial L}{\partial \mathbb{W}},
\end{equation}
where $\theta$ is the step size of learning rate. Furthermore, in weighted cross-entropy, adjusting the weight $q(\alpha)$ for the loss of minor class is essentially regulating its impact on the learning rate.}

{According to \eqref{single_pixel_wce}, the loss function decays to cross-entropy when $q(\alpha)=1$ holds; the loss function becomes the vanilla weighted cross-entropy used by Xie et al.\cite{xie2015holistically} when $q(\alpha)=\alpha/(1-\alpha)$ holds. The goal of machine learning, including deep learning, is to minimize the loss of the model. In the training process of the model, if the learning rate is too small, the loss will easily fall into a local minimum; if the learning rate is too large, the loss will easily skip the minimum value and oscillate back and forth. Note that `easily' mentioned here refers to a high probability of occurrence during the entire training process of the model. Meanwhile, in our WCE, the proposed weight is between the two, i.e., $q(\alpha) \in [1, \alpha/(1-\alpha)]$. {\emph{According to the formulas \eqref{pde3} and \eqref{update rule}, applying the weighted method we proposed is equivalent to utilizing a \textbf{moderate} learning rate to update the network parameters during training.}}}

{To validate that our weighted system corresponds to a moderate learning rate, we make a further elaboration by combining with \reffig{The validity of the loss function}. When $q(\alpha)=1$ holds, the training loss appears non phenomena and we do not display the corresponding experimental results. Note that in footnote\footref{footnote_4}, as the fine-tuning coefficient $\beta$ becomes smaller and then is less than a certain threshold, the corresponding weight $q$ is closer to 1 and next the loss function appear non phenomena because the corresponding learning rate become too small.
When $q(\alpha)=\alpha/(1-\alpha)$ holds (corresponding to pink line in \reffig{The validity of the loss function}), the metric results fluctuate in many training epochs because the corresponding learning rate is relatively large. While utilizing our weighted method (corresponding to blue line in \reffig{The validity of the loss function}), although the metric results have three slightly large shocks, they show an upward trend overall with the increase of training epochs.}

\subsection{Visual Comparison on Other Databases}\label{Appendix B}
Considering the limited main-body space, we put the visualization of the other three databases in this section.

\begin{figure*}[htbp]
    \centering
    \includegraphics[width=16.8cm]{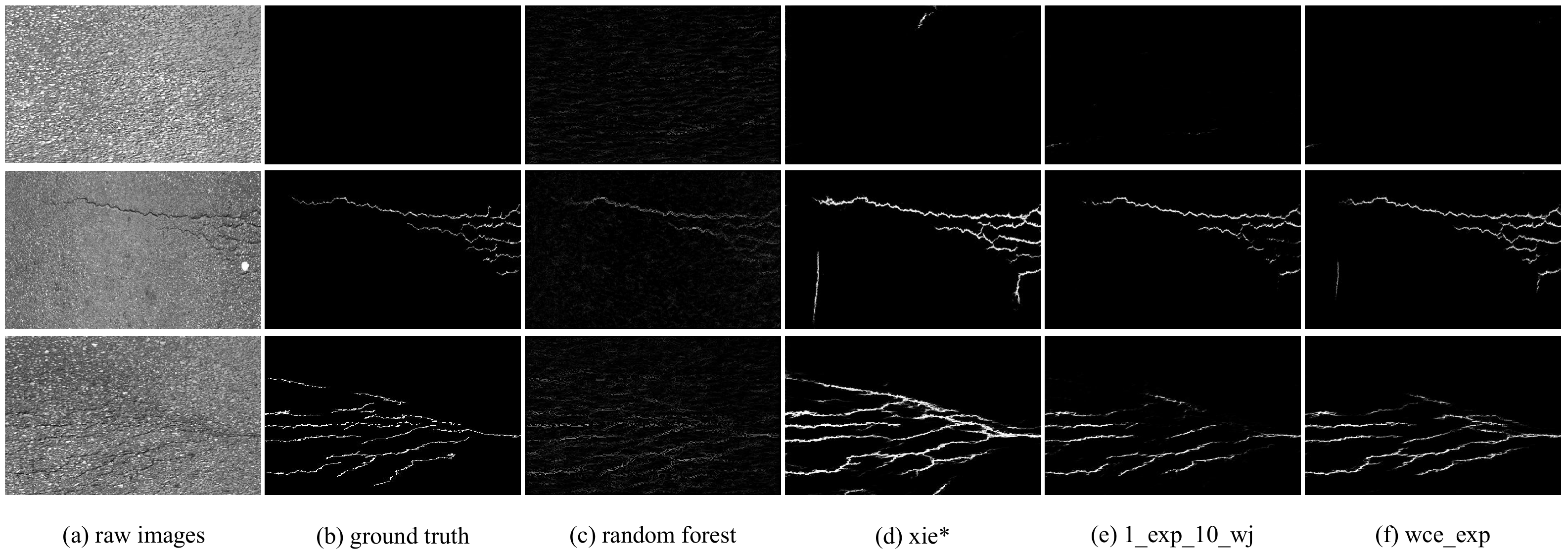}
    \caption{Qualitative examples of using different methods on AigleRN. (a) contains the raw images; (b) corresponds to the {ground truth}; (c) and (d) are the results obtained by the first and second baselines; (e$\sim$f) correspond to the ones originated from our loss functions with 4 and 15 training epochs, respectively. Note that the second baseline is with 50 training epochs. These methods are mentioned in \reftab{The fast and accurate results on AiglRN database}, respectively.}
    \label{visual comparision on AiglRN}
\end{figure*}

\begin{figure*}[htbp]
    \centering
    \includegraphics[width=16.8cm]{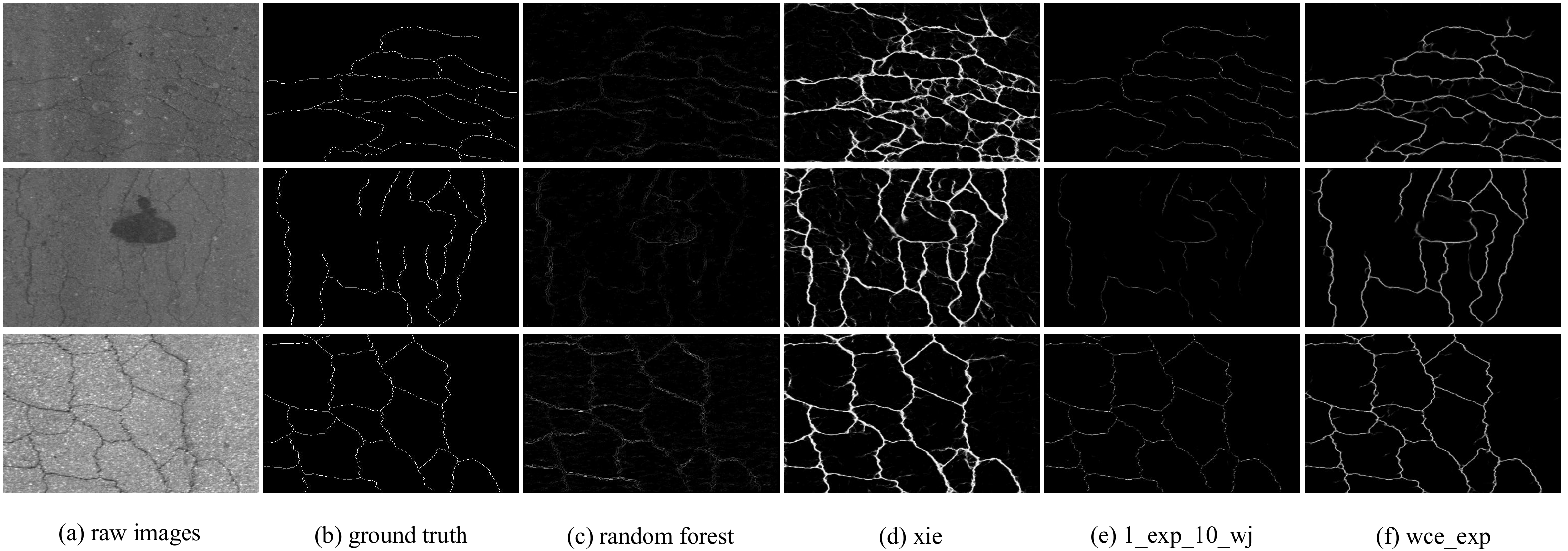}
    \caption{Qualitative examples of using different methods on Crack360. (a) contains the raw images; (b) corresponds to the {ground truth}; (c) and (d) correspond to the results obtained by the first and second baselines; (e$\sim$f) are the ones originated from our loss functions with 3 and 10 training epochs, respectively. Note that the second baseline is with 70 training epochs.}
    \label{visual comparision on Crack360}
\end{figure*}

\begin{figure*}[htbp]
    \centering
    \includegraphics[width=16.8cm]{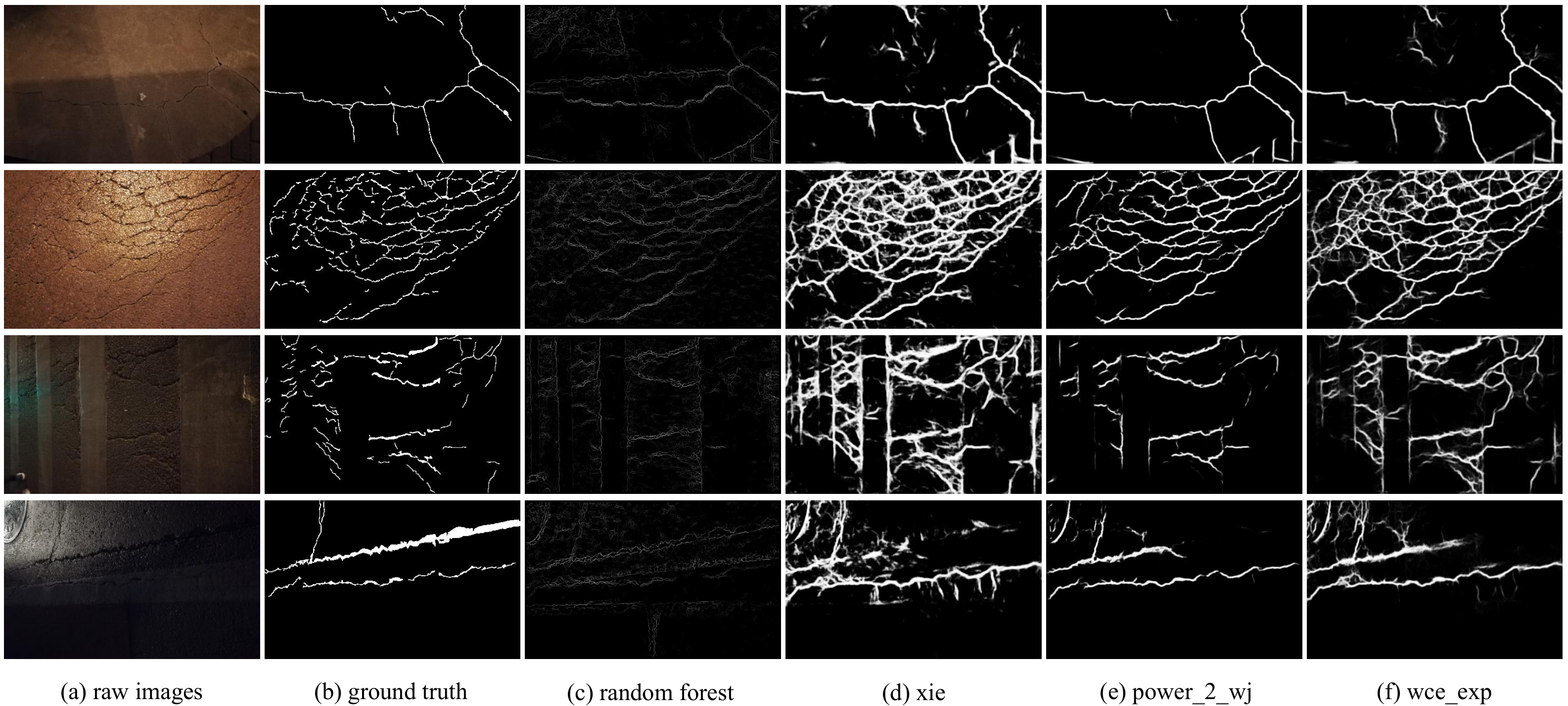}
    \caption{Qualitative examples of using different methods on BJN260. (a) contains the raw images; (b) corresponds to the {ground truth}; (c) and (d) are the results obtained by the first and second benchmarks; (e$\sim$f) correspond to the results originated from our loss functions with 5 and 15 training epochs, respectively. Note that the second baseline is with 30 training epochs.}
    \label{visual comparision on BJN260}
\end{figure*}

\end{document}